\documentclass[10pt,journal,compsoc]{IEEEtran}
\usepackage{color}
\usepackage{url,hyperref}
\usepackage{graphicx}
\usepackage{amsmath}
\usepackage{graphicx}
\usepackage{times}
\usepackage{epsfig}
\usepackage{url}
\usepackage{caption}
\usepackage{subcaption}
\usepackage{amstext}
\usepackage{amssymb}
\usepackage{amsthm}
\usepackage{color}
\usepackage{amsfonts}
\usepackage{paralist}
\usepackage{algorithm}
\usepackage{algorithmic}
\usepackage{pseudocode}
\usepackage{latexsym}
\usepackage{wasysym}
\usepackage{harmony}
\usepackage{gensymb}
\usepackage{textcomp}
\usepackage{threeparttable}
\usepackage{booktabs}
\usepackage{array}
\usepackage{multirow}
\usepackage{enumitem}
\usepackage{changes}
\definechangesauthor[name={Vlad}, color=orange]{Vlad}
\definechangesauthor[name={Hanhe}, color=blue]{Hanhe}
\definechangesauthor[name={Tamas}, color=orange]{Tamas}
\definechangesauthor[name={Dietmar}, color=red]{Dietmar}

\ifCLASSINFOpdf
\else
\fi
\begin{document}
%

\title{KonIQ-10k: An ecologically valid database for deep learning of blind image quality assessment}


\author{Vlad Hosu$^{a,c}$, Hanhe Lin$^{a,c}$, Tamas Sziranyi$^{b}$, and Dietmar Saupe$^a$
	\thanks{$^a$Department of Computer and Information Science, University of Konstanz, 78464 Konstanz, Germany.
E-mail: \{vlad.hosu, hanhe.lin, dietmar.saupe\}@uni-konstanz.de}
    \thanks{$^b$Machine Perception Research Laboratory, Institute for Computer Science and Control (SZTAKI), 1111 Budapest, Hungary.
E-mail: sziranyi@sztaki.hu}
	\thanks{$^c$Vlad Hosu and Hanhe Lin contributed equally.}
	\thanks{This paper has supplementary downloadable material available at http://ieeexplore.ieee.org., provided by the author. This includes the "Supplementary material" PDF. Contact vlad.hosu@uni-konstanz.de or hanhe.lin@uni-konstanz.de for further questions about this work.}
}

\markboth{IEEE TRANSACTIONS ON IMAGE PROCESSING,~Vol.~29, pp.~4041-4056, ~2020}%
{Shell \MakeLowercase{\textit{et al.}}: Bare Demo of IEEEtran.cls for IEEE Journals}
%



\maketitle

\begin{abstract}
Deep learning methods for image quality assessment (IQA) are limited due to the small size of existing datasets. Extensive datasets require substantial resources both for generating publishable content and annotating it accurately.
We present a systematic and scalable approach to creating KonIQ-10k, the largest IQA dataset to date, consisting of 10,073 quality scored images. It is the first in-the-wild database aiming for ecological validity, concerning the authenticity of distortions, the diversity of content, and quality-related indicators. Through the use of crowdsourcing, we obtained 1.2 million reliable quality ratings from 1,459 crowd workers, paving the way for more general IQA models.
We propose a novel, deep learning model (KonCept512), to show an excellent generalization beyond the test set ($0.921$ SROCC), to the current state-of-the-art database LIVE-in-the-Wild ($0.825$ SROCC). The model derives its core performance from the InceptionResNet architecture, being trained at a higher resolution than previous models ($512\times384$). Correlation analysis shows that KonCept512 performs similar to having $9$ subjective scores for each test image.

\end{abstract}
\begin{IEEEkeywords}
Image database, diversity sampling, crowdsourcing, blind image quality assessment, subjective image quality assessment, convolutional neural networks, deep learning
\end{IEEEkeywords}

%
\IEEEpeerreviewmaketitle

\section{Introduction}
%
%
Image Quality Assessment (IQA) plays an essential role in a broad range of applications ranging from image compression to machine vision, and more \cite{wang2011applications,Kopilovic_2005_Sziranyi, dodge2016understanding,men2019visual}. 
Ideally, the visual quality of images is assessed by subjective user studies involving experts in a controlled environment to yield Mean Opinion Scores (MOS). The MOS is a direct measure of the perceived quality of images, which is important both for choosing the right technology and for making further improvements to existing imaging technologies. However, subjective studies are time-consuming and expensive and have limited applicability in practice. Hence, objective IQA,  
i.e., algorithmic estimation of visual quality has been a long-standing research topic, which has recently attracted more attention.

According to the availability of pristine reference images, objective IQA methods are categorized as Full-Reference (FR), e.g., SSIM \cite{wang2004image}, 
Reduced-Reference (RR), e.g., RRED \cite{Soundararajan:2012}, and No-Reference (NR), e.g., CORNIA \cite{ye2012unsupervised},  HOSA \cite{hosa}. In comparison to FR and RR, NR or Blind IQA (BIQA) is the more challenging problem among the three. BIQA is also the most practical of the three since reference-based comparisons are not available in many applications.

The development of objective IQA methods requires data\-bases that provide images and carefully gathered subjective quality scores. Such databases are also required to improve existing methods via benchmarking and to provide a source of data for training and parameter fitting new models. In order for the trained models to be generally applicable, they need to use representative data from the real world, in terms of authenticity, scale, and diversity. The ecological validity of an IQA database refers to the representativeness of the visual collection for a wide range of real-world photos such as those available on the Internet, and the generalization potential of models trained on it. These are both important goals of our work.

Conventionally, creating an IQA database has followed a standard procedure: collect pristine images and artificially degrade them. Then a few volunteers, usually naive participants, would be asked to assess the quality of the distorted images. The first drawback of this approach is that the diversity of image content is limited since all the distorted images are degraded from a small set of pristine images. Second, the distortions are applied in very limited combinations, whereas ecologically valid distortions, such as those of public Internet images, are caused by combinations of distortions of types that differ from those in the databases. For instance, some authentic degradation types such as wrong focus or motion blur due to object movement are hard to reproduce artificially by distorting a pristine image. Last, but not least, the conventional approach for creating IQA datasets results in smaller databases, since assessing the quality of a large number of images in a lab setting is too costly. Nonetheless, lab studies are usually well controlled and produce strongly consistent opinions.

Recently, deep learning has achieved promising results in a number of computer vision tasks: image classification \cite{vgg} \cite{he2016deep}, object detection \cite{ren2015faster} \cite{redmon2016you},  and BIQA \cite{boss} \cite{conv1} \cite{bianco2018use}. It is widely believed that a large-scale IQA database, diverse in content and authentic in distortions, could benefit the development and evaluation of deep learning approaches. However, all existing deep-learning-based BIQA methods are trained and evaluated on small and artificially distorted IQA databases. Assessing the quality of a very large number of images in a lab setting would require too many participants and too much time for preparing and running the experiment.



In order to address the limitations of existing databases, our work provides the following significant contributions:
\begin{itemize}[leftmargin=*]
\item We designed an approach easily scaled, which allowed us to create the largest IQA database to date (KonIQ-10k), concerning the number of images and subjective scores:
    \begin{itemize}[leftmargin=*]
    \item Consisting of 10,073 images, selected from 10 million YFCC100M \cite{thomee:2016} entries. Our sampling algorithm ensures the diversity of content and distortions, making use of seven indicators for quality and one for content based on deep features.
    \item For each image, 120 reliable quality ratings were obtained by crowdsourcing, performed by 1,459 crowd workers.
    \end{itemize}  
\item We proposed an end-to-end deep-learning-based BIQA method:
    \begin{itemize}[leftmargin=*]
    \item It is a unified transfer learning approach based on fine-tuning a pre-trained  Convolutional Neural Network (CNN).
    \item With an unified architecture, compared the performance of five state-of-the-art CNNs.
    \item Compared the performance of five loss functions governing the criteria by which the subjective scores are predicted, where two of them were used to predict MOS directly, and the other three were applied to predict distributions of ratings.
    \item Explored the effect of the training set size on the performance of the proposed best model -- KonCept512.
    \item Showed that KonCept512, trained on KonIQ-10k, works well on another IQA database, all through cross-database testing.
    \end{itemize}
\end{itemize}

\section{Related work}

\begin{table*}[ht]
\setlength\extrarowheight{2pt}
\footnotesize
\caption{Comparison of existing IQA databases with KonIQ-10k.}
\label{tb:benchmarkdb}
\centering
\resizebox{1.0\textwidth}{!}{
\begin{tabular}{l c r r r c r r c} \hline
& & &\multicolumn{1}{c}{No.\ of}&& No.\ of&\multicolumn{1}{c}{No.\ of}&\multicolumn{1}{c}{Ratings}  \\
Database & Year &Content&distorted images &Distortion type&distortion types &rated images &per image &Environment\\ \hline
IVC \cite{ivcdb}& 2005&10&185& artificial&4&185&15&lab\\
LIVE \cite{sheikh:2006statistical} & 2006 &29&779& artificial & 5&779&23&lab \\
TID2008 \cite{ponomarenko:2009tid2008}&2009&25&1,700& artificial &17&1,700&33&lab \\
CSIQ \cite{larson:2010most} & 2009&30 &866&artificial& 6&866&5$\sim$7&lab \\
TID2013 \cite{ponomarenko:2015image}&2013&25&3,000&artificial&24&3,000&9&lab \\
CID2013 \cite{virtanen2015cid2013}&2013&8&474&authentic&12$\sim$14&480&31&lab \\
LIVE-itW \cite{ghadiyaram:2016massive}& 2016&1,169&1,169&authentic&N/A&1,169&175&crowdsourcing \\
Waterloo Exploration \cite{ma2016group}& 2016& 4,744& 94,880 & artificial& 4 & 0&0 & lab \\
MDID \cite{sun2017mdid} & 2017 & 20 &1,600 & artificial &5 & 1,600 & 33$\sim$35 & lab \\
KADID-10k \cite{lin2019kadid} & 2019 & 81 & 10,125 & artificial & 25 & 10,125 & 30 & crowdsourcing \\
\hline
 KonIQ-10k&2018&10,073&10,073&authentic&N/A&10,073&120&crowdsourcing \\
\hline
\end{tabular}
}
\end{table*}

The related work is divided into two parts according to the contributions of the article. After briefly discussing the literature concerning the creation of benchmark IQA databases, we review state-of-the-art blind IQA methods.

\subsection{Creating benchmark IQA databases}
A number of IQA databases have been released in recent years, aiming to help the development and evaluation of objective IQA methods, see Table~\ref{tb:benchmarkdb}. 

The early, conventionally built IQA database IVC \cite{ivcdb} was released in 2005. LIVE \cite{sheikh:2006statistical}, TID2008 \cite{ponomarenko:2009tid2008}, and CSIQ \cite{larson:2010most} were the most commonly used to develop, improve, and evaluate objective IQA methods. TID2008 was further extended into TID2013 \cite{ponomarenko:2015image} by including seven more distortion types. 
Rather than exclusively degrading each reference image with one type of distortion, the distorted images in MDID \cite{sun2017mdid} were degraded by multiple distortions of random types and levels. The databases above were small-scaled, thus were unable to benefit the development of deep-learning-based IQA methods. KADID-10k \cite{lin2019kadid} was proposed to address the limitation with 10,125 distorted images. However, such a database still contains minimal content types (81 reference images) and a few types of artificial distortions (25 distortions).

Virtanen et al.\ \cite{virtanen2015cid2013} were the first to introduce more authentic distortions, creating 480 images of 8 different scenes captured by 79 different cameras. However, the creation method was time-consuming and expensive and thus impractical for large-scale databases. Ghadiyaram et al.\ \cite{ghadiyaram:2016massive} created LIVE in the Wild (LIVE-itW) by asking a few photographers to capture 1,169 images by a variety of mobile cameras. Their visual quality was assessed by crowdsourcing experiments. Although this method provided an alternative way to reduce the time and cost for the IQA subjective study, the database size, as well as the content diversity, were still relatively low, having near-duplicate photos -- w.r.t.\ content -- that were captured from the same scenes.

Ma et al.\ \cite{ma2016group} created a database with 4,744 pristine images and 94,880 distorted images to validate their proposed mechanism called group MAximum Differentiation (gMAD) competition. Their database was meant to provide an alternative evaluation for the performance of IQA models by means of paired comparisons. Although the Waterloo Exploration database was the largest available in the field, its images were artificially distorted, thus unauthentic, and due to the lack of subjective ratings, it could not easily be used for developing new IQA methods that rely on them.

In comparison to lab-based IQA studies that are limited in size, crowdsourcing has been successfully applied for larger databases of images \cite{ghadiyaram:2016massive} and videos \cite{hosu:2017konstanz} (to about 1,000 stimuli).  Although it was believed that data collected by crowdsourcing was less reliable, 
Siahaan et al.\ \cite{siahaan_reliable_2016} and other studies \cite{QoMEXReliability} verified that crowd workers could generate reliable results under specific experimental setups. 

Taking a different approach to traditional rating datasets, Yu et al.\ \cite{yu2018learning}  built an extensive, 12,853 natural images database annotating the severity of seven perceptual defect types via crowdsourcing. The authors collected a minimal number of ratings for each defect (five), resulting in less precise subjective scores. This database does not contain traditional quality scores, so we did not include it in Table \ref{tb:benchmarkdb}.

\subsection{Blind image quality assessment methods}
\label{sec:RelatedWorkBIQA}

In terms of the methodology used, we can categorize BIQA  into conventional and deep-learning-based methods.

\subsubsection{Conventional BIQA methods}
Similar to conventional image processing and computer vision tasks, conventional BIQA methods require domain experts to engineer and design a feature extractor carefully. The feature extractor will transform raw image data into a representative feature vector from which a regression model, e.g., Support Vector Regression (SVR), 
can predict the MOS. 

Many conventional BIQA methods were derived from the Natural Scene Statistics (NSS) model \cite{srivastava2003advances}. Such methods seek to capture the natural statistical behavior of images. Moorthy et al.\ proposed the BIQI \cite{Moorthy:2010} method, in which a description of Distorted Image Statistics (DIS) captures the NSS changes resulting from distortions. With DIS features, the method has a two-step framework: image distortion classification followed by a distortion-specific quality assessment. With the same two-step framework, two more methods, DIIVINE \cite{Moorthy:2011} and SSEQ \cite{Liu:2014b}, were proposed to improve the performance of quality assessment.
BLIINDS-II \cite{blind2} relies on a Bayesian inference model to predict MOS based on certain extracted features which were derived using an NSS model of the image's discrete cosine transform coefficients. A simple but efficient method called BRISQUE, was proposed in \cite{bris}. Using scene statistics of locally normalized luminance coefficients, it quantified possible losses of naturalness in the image due to the presence of distortions.


Another direction for conventional BIQA methods is the Bag-of-Words (BoW) model that uses local features. Ye et al.\ \cite{ye2012unsupervised} proposed an unsupervised feature learning approach named CORNIA. It learned a dictionary by clustering raw image patches extracted from a set of unlabeled images. An image was represented with a histogram for quality assessment by softly assigning raw image patches to the dictionary with max pooling. Following the same idea, HOSA was proposed in \cite{hosa}. Apart from softly assigning patches to the corresponding means of each cluster, HOSA softly aggregates the differences of high order statistics (mean, variance, and the skew present) between raw image patches and corresponding clusters. This global quality-aware image representation reduced computational costs and improved performance.

%
%

%
%
%
%



\subsubsection{Deep-learning-based BIQA methods}
Instead of carefully designing handcrafted features, deep-learning-based BIQA methods strive to automatically discover representations from raw image data that are most suitable for predicting quality scores.

Due to the small size of existing IQA databases unsuitable for end-to-end learning, several BIQA methods applied deep learning methods as feature extractors. Ghadiyaram et al. \cite{deepbelief} used an unsupervised deep neural network, i.e., deep belief nets, to discover representative features for quality prediction. In \cite{bianco2018use}, a VGG16 network was fine-tuned as a feature extractor. Their best method was called DeepBIQ, and it predicts image quality by average-pooling the predicted MOS on multiple image patches, where the score of each patch is determined by training an SVR.

Similarly, the BLINDER model was proposed by Gao et al.\ \cite{gao2018blind}. By feeding an image into a VGG16 network and generating one feature vector in each layer, a quality score was created and then estimated for each feature vector by SVR. The overall quality of the image was the mean of these level-wise predicted quality scores. 

To address the limitation of small training data and to implement end-to-end learning, an alternative way was introduced to train on sampled image patches instead of entire images. The assumption made was that a set of sampled image patches has the same quality score as the entire image. The quality of an image was estimated by averaging the predicted quality scores of sampled image patches. This idea was implemented in \cite{conv1} and \cite{boss} by training a CNN with $32 \times 32$ RGB image patches. The CNN model in \cite{boss} is deeper, having 12 layers, compared to the seven layers in \cite{conv1}. Training on image patches allowed it to train a deep CNN from scratch. However, it ignored content information of an image, whereas IQA had been shown to be content-dependent \cite{siahaan2016does}.

Transfer learning was also applied to address the small data issue. 
Liu et al.\ proposed the RankIQA method \cite{Liu_2017_ICCV}. By generating pairs of pristine and artificially distorted images (for which the quality ordering was known), the authors trained a Siamese Network to rank the quality of image pairs. The quality score of a single image was estimated by fine-tuning a network that relied on the knowledge represented (its features) in the Siamese Network. Ma et al.\ \cite{ma2018end} proposed the MEON model. It was a multi-task model consisting of two sub-networks, a distortion identification network, and a quality prediction network, where both the networks shared their weights in early layers. Since it is easy to generate training data synthetically, it trained the distortion identification sub-network to have pre-trained weights in the first layers. With the pre-trained early layers and the outputs of the first sub-network, training a quality prediction sub-network on the small IQA databases became feasible.
Transfer learning is not limited to blind IQA, Prabhushankar et al.~\cite{prabhushankar2017ms} proposed MS-UNIQUE, an FR-IQA method. The quality feature vector of a given image was represented by multiple linear decoders, which had been previously trained on 100,000 image patches from ImageNet. The visual quality was estimated by comparing the feature vectors corresponding to the original and distorted images.

Talebi et al.\ \cite{talebi_nima_2017} introduced an end-to-end trained framework that relies on existing object-classification-architectures such as MobileNet, VGG16, and Inception-v2, with an added simple regression head to predict the distribution of scores for either aesthetics and image quality assessment. They used two fully-connected layers on top of the base network, thus limiting the application of their approach to images with a fixed size of $224\times224$ pixels. They used Stochastic Gradient Descent with very low learning rates in order to avoid early over-fitting. We consider their suggestions for our framework, and further expand on them.

Dendi et al. \cite{dendi2018generating} used a convolutional auto-encoder for distortion map generation. The training ground-truth distortion map was estimated by a well-known FR-IQA measure, i.e., SSIM. The authors show that the performance of conventional BIQA methods can be improved by predicting distortion maps.

Zhang et al. \cite{zhang2019learning} used a Siamese network to learn to rank image pairs. The ground-truth binary labels for the Siamese network are obtained from the corresponding image MOS. The absolute quality score for an image is reconstructed from the pair-wise predictions using the Bradley-Terry model.

\section{Database creation}
\label{sec:databcreate}

The primary use of our IQA database is training better deep learning models. Existing IQA databases are either too small or rely on a limited set of pristine images and artificial, non-authentic quality degradations. Since IQA is content-dependent and authentic image degradations are essential for accurate quality predictions, we created an extensive collection of images in-the-wild to depict a broad range of appropriate content and authentic distortions.

To build a balanced dataset, we relied on multiple indicators that characterize image quality. Indicator values in-the-wild are predominantly imbalanced. Vonikakis et al.~\cite{vonikakis2016shaping} have argued for creating better datasets via re-targeting the distribution of multiple indicators, such that all values along each dimension are more equally represented. In a similar vein, Hosu et al.~\cite{hosu:2017konstanz} have employed quality-related indicators that are used to create diverse datasets by "fair-sampling", another way of describing the balancing process.

\subsection{Overview}

Our goal of creating an authentically distorted database was achieved by selecting images from a massive public multimedia database, YFCC\-100m \cite{thomee:2016}. We randomly selected approximately ten million (9,974,030) image records. Then we filtered them down in two stages obtaining the final database.

\begin{figure}[!t]
\centering
\vspace{-10pt}
\includegraphics[width=0.4\textwidth]{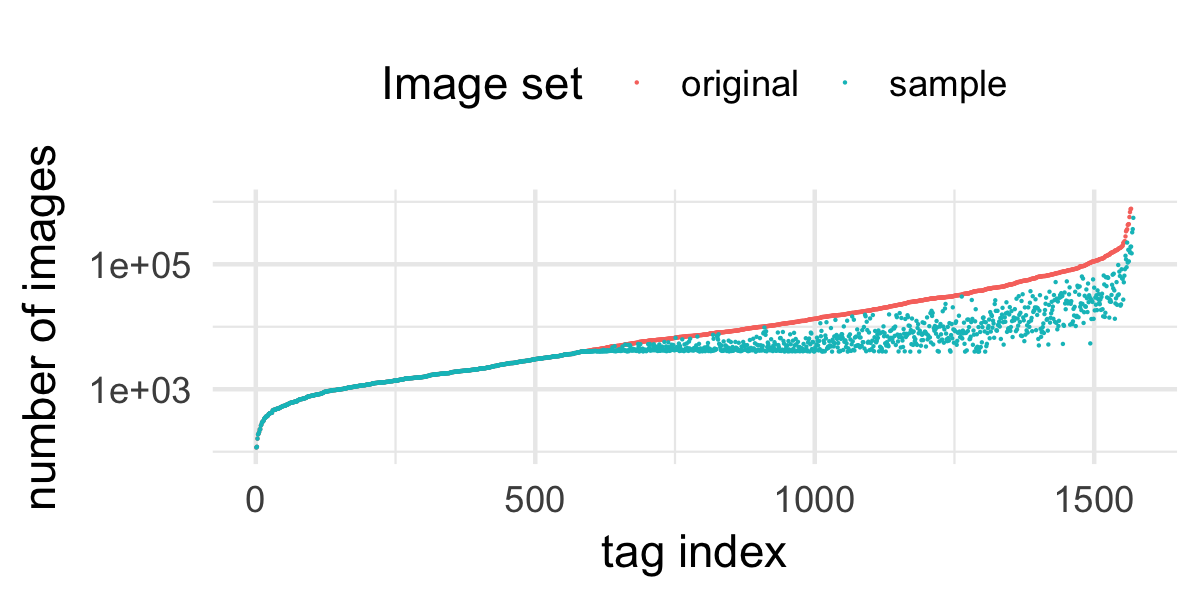}
\caption{Sampling 1.0 from 4.8 million images. The tags were sorted according to increasing frequency in the pre-sample set (red). The histogram of the number of sampled images per tag is shown in cyan. The two histograms start diverging at the quota $Q=4000$ images per tag. Ideally, the sampled histogram (cyan) should be flat after the divergence point, however this cannot be achieved as each image can have multiple tags.\vspace{-5pt}}
\label{fig:tag_sampling}
\vspace{-5pt}
\end{figure}

In the first stage, we selected images with an appropriate Creative Commons (CC) license that allows editing and redistribution and chose those with available machine tags (from YFCC100m) and a resolution between $960\times540$ and $6000\times6000$. From this set of 4,807,816 images, we proposed a new tag-based sampling procedure that was used to select one million images such that their machine tag distribution covers the larger set well, see Fig.~\ref{fig:tag_sampling}.

In the second stage, all images in the set of one million that were larger than  $1024\times768$ were downloaded and re-scaled to $1024\times768$ pixels, and cropping was applied to maintain the pixel-aspect ratio. In order to keep the faces' salient parts of the image in the frame, we designed our own cropping method. It relies on the Viola-Jones face detector and the saliency method of Hou et al.\ \cite{hou_image_2012}. 13,000 images were then sampled while enforcing a uniform distribution across eight image indicators. Duplicates were removed, using a sampling strategy that accounts for category and indicators. This collection was manually filtered for inappropriate content resulting in our KonIQ-10k\footnote{The database and code is available at \url{http://database.mmsp-kn.de}} dataset of 10,073 images, slightly above our target size that we had set to 10,000.

\subsection{Initial tag-based content sampling}
\label{sec:tagsampling}
Downloading 4.8 million images consumes significant bandwidth and storage space. Hence, we devised a way to shrink the set from one million images to maintain content diversity. We aimed to achieve a  "uniform" coverage of tags by sampling at least a certain number of images for each (a quota). This is generally not precisely possible, as images have more than one tag (9.2 on average). Therefore, we devised a simple and computationally efficient sampling heuristic while keeping the above objectives in mind. The heuristic chooses images filling the given quota for less popular tags first until the total number of required images is selected. See the appendix for more details. The distribution of images by tag is shown in Fig. \ref{fig:tag_sampling}.

\begin{figure*}[t]
\centering
\begin{minipage}{0.119\linewidth}
\centerline{\includegraphics[width=\textwidth]{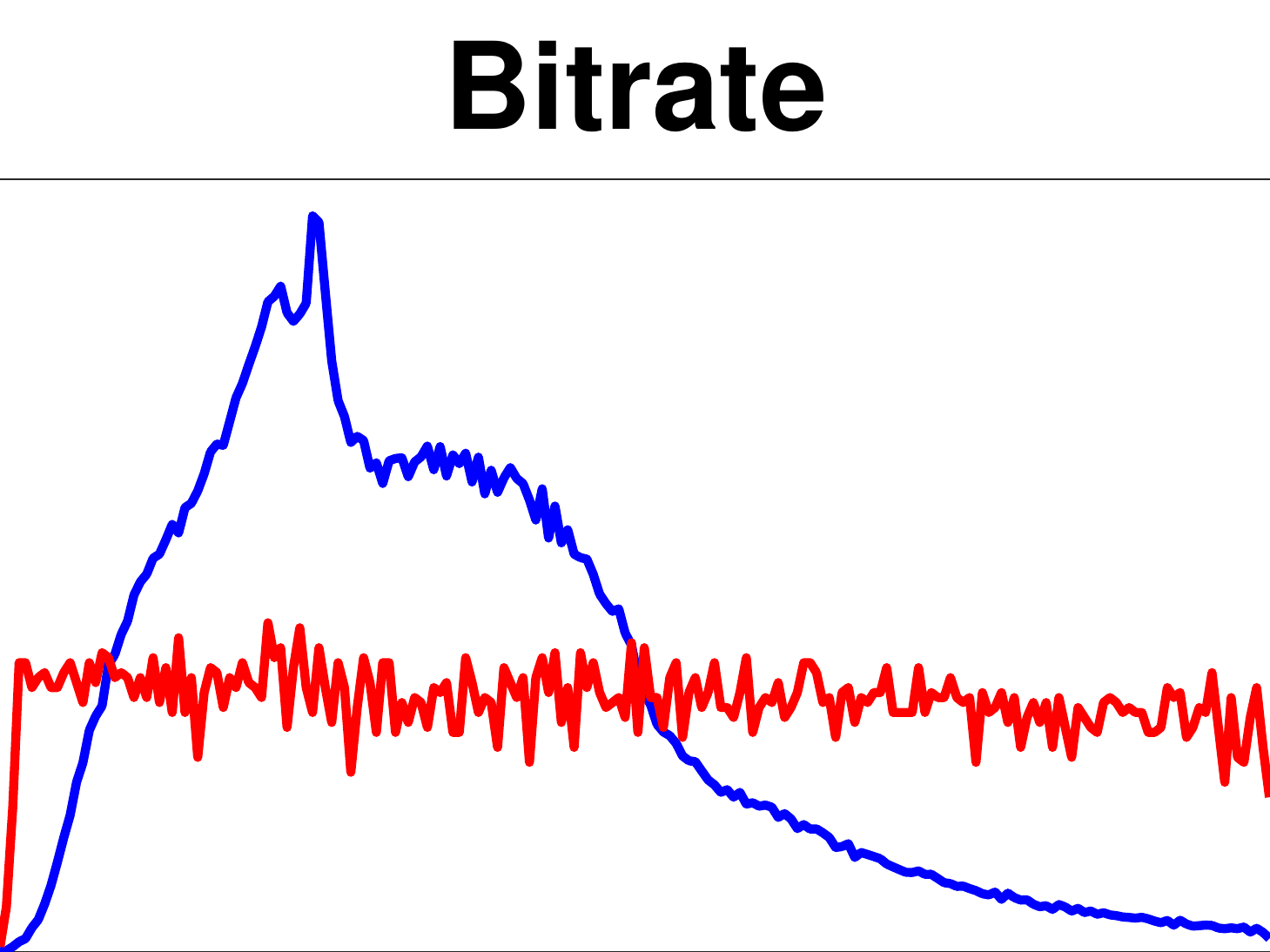}}
\end{minipage}
\begin{minipage}{0.119\linewidth}
\centerline{\includegraphics[width=\textwidth]{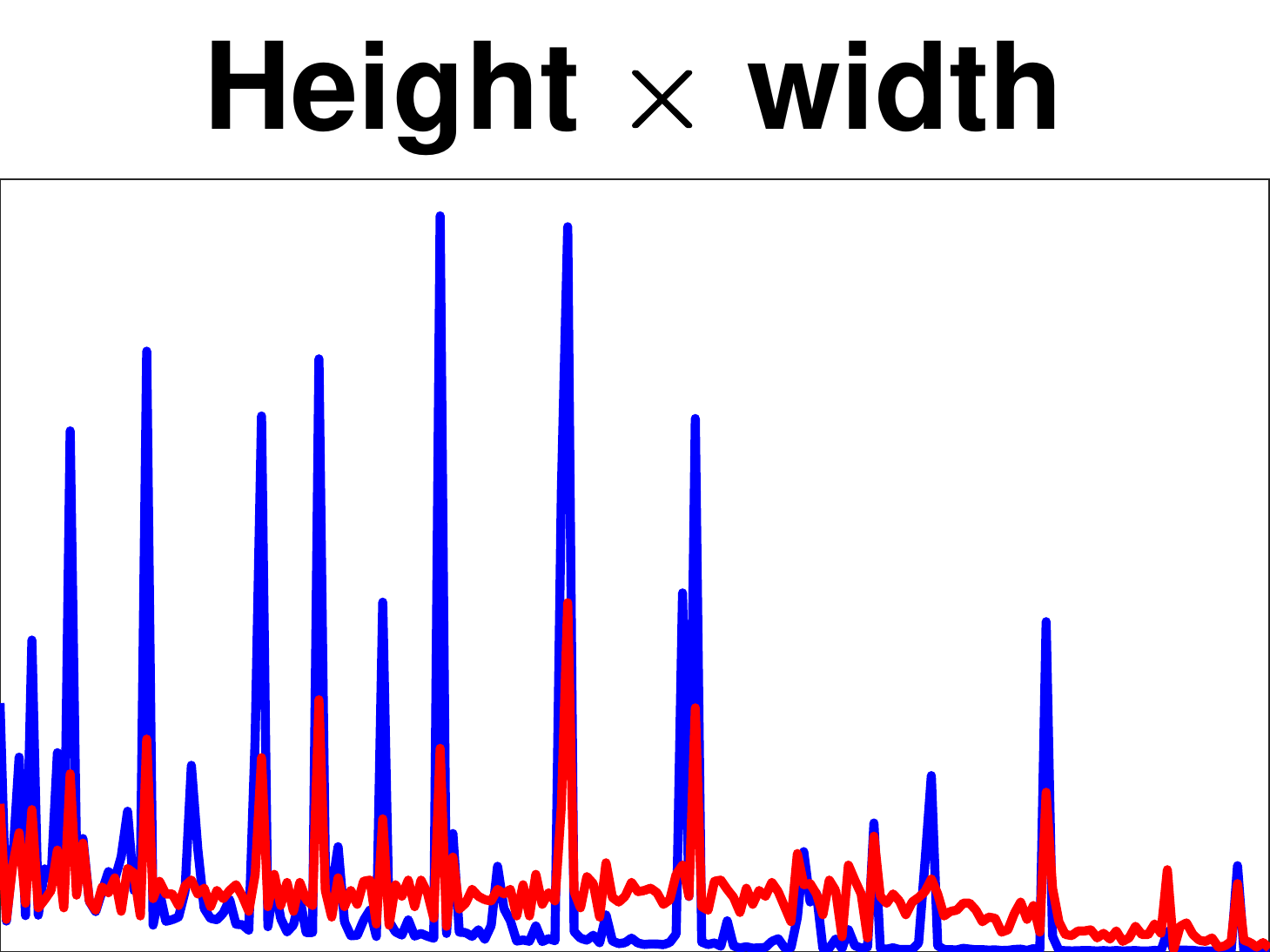}}
\end{minipage}
\begin{minipage}{0.119\linewidth}
\centerline{\includegraphics[width=\textwidth]{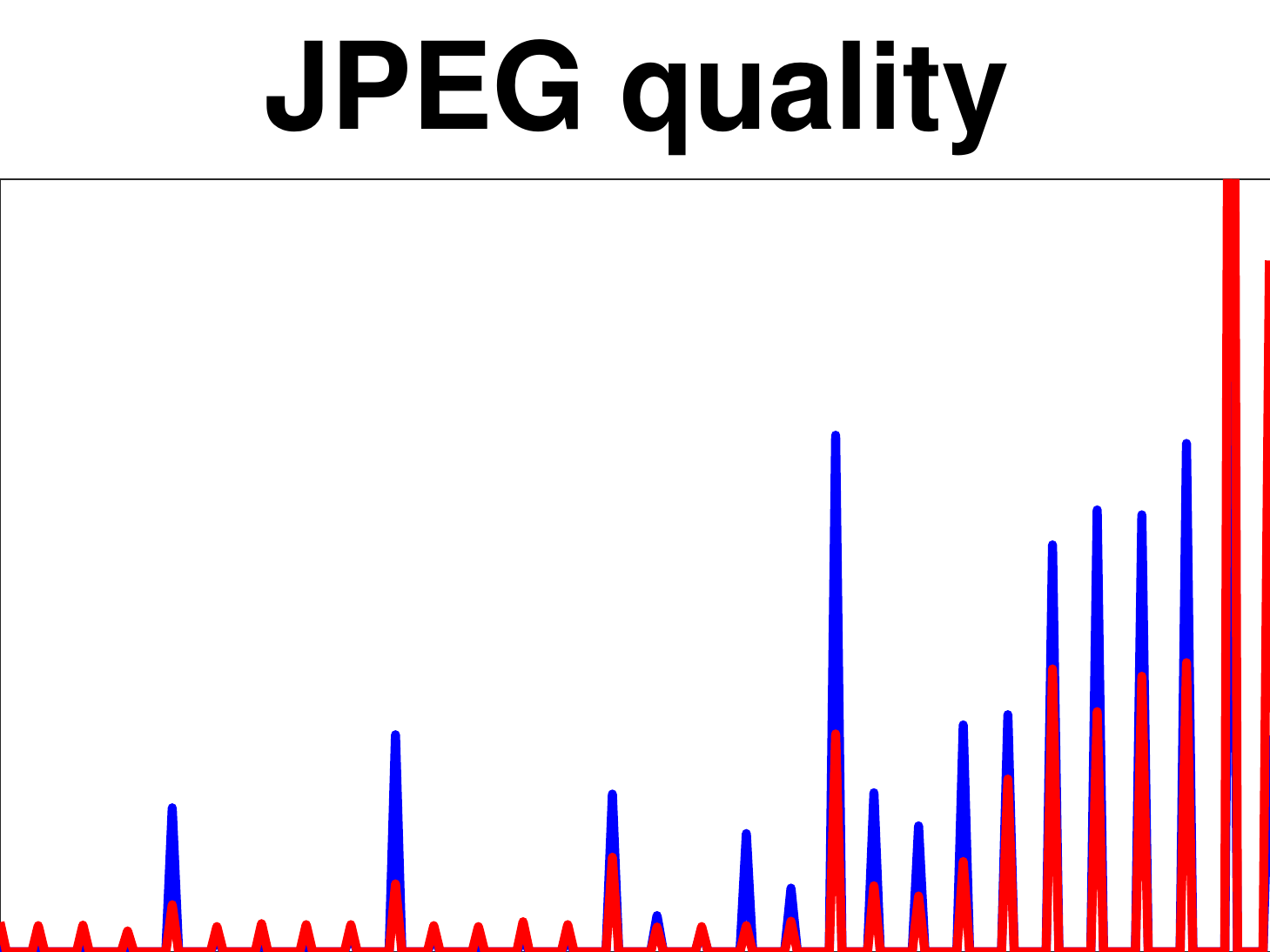}}
\end{minipage}
\begin{minipage}{0.119\linewidth}
\centerline{\includegraphics[width=\textwidth]{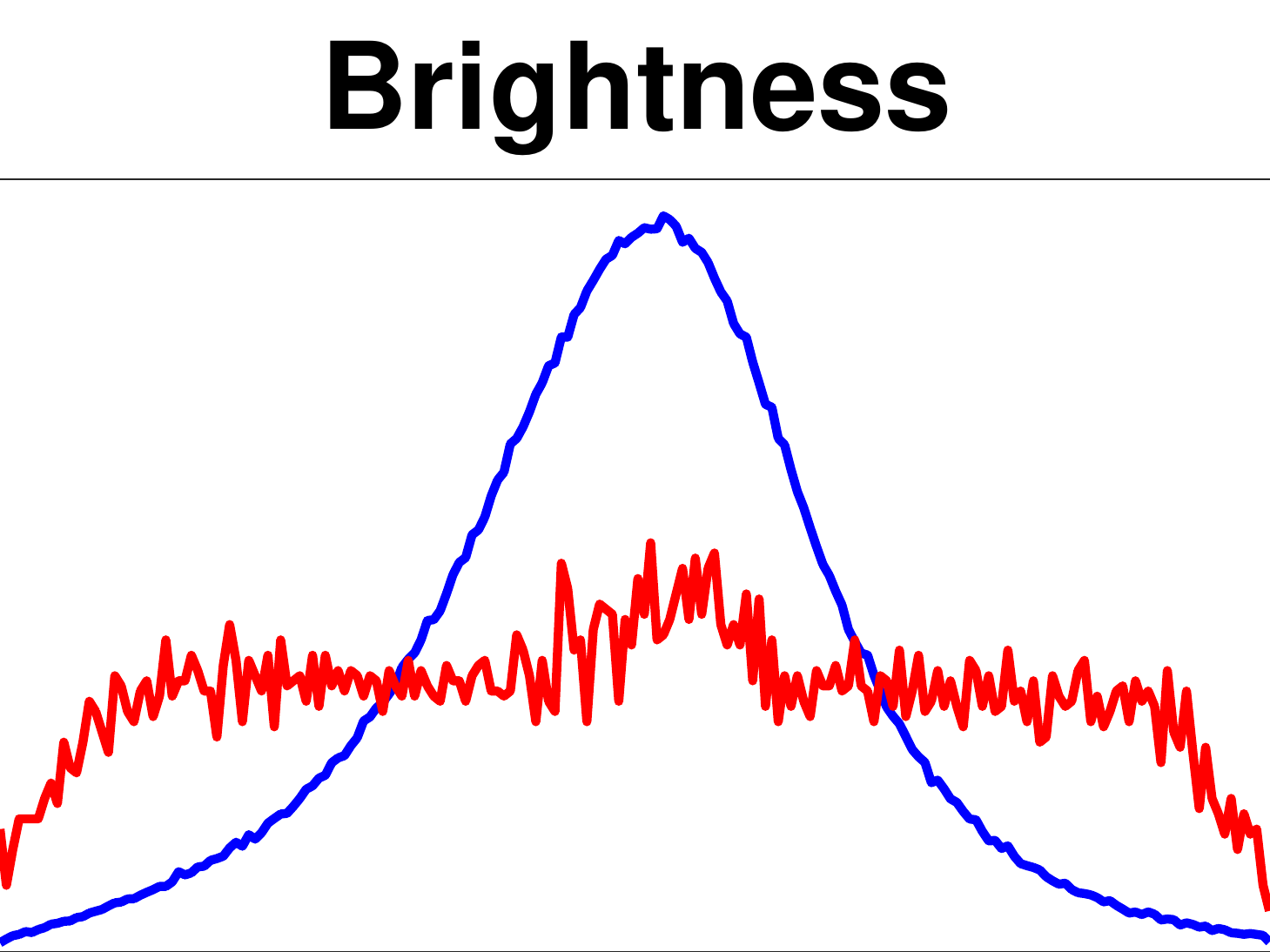}}
\end{minipage}
\begin{minipage}{0.119\linewidth}
\centerline{\includegraphics[width=\textwidth]{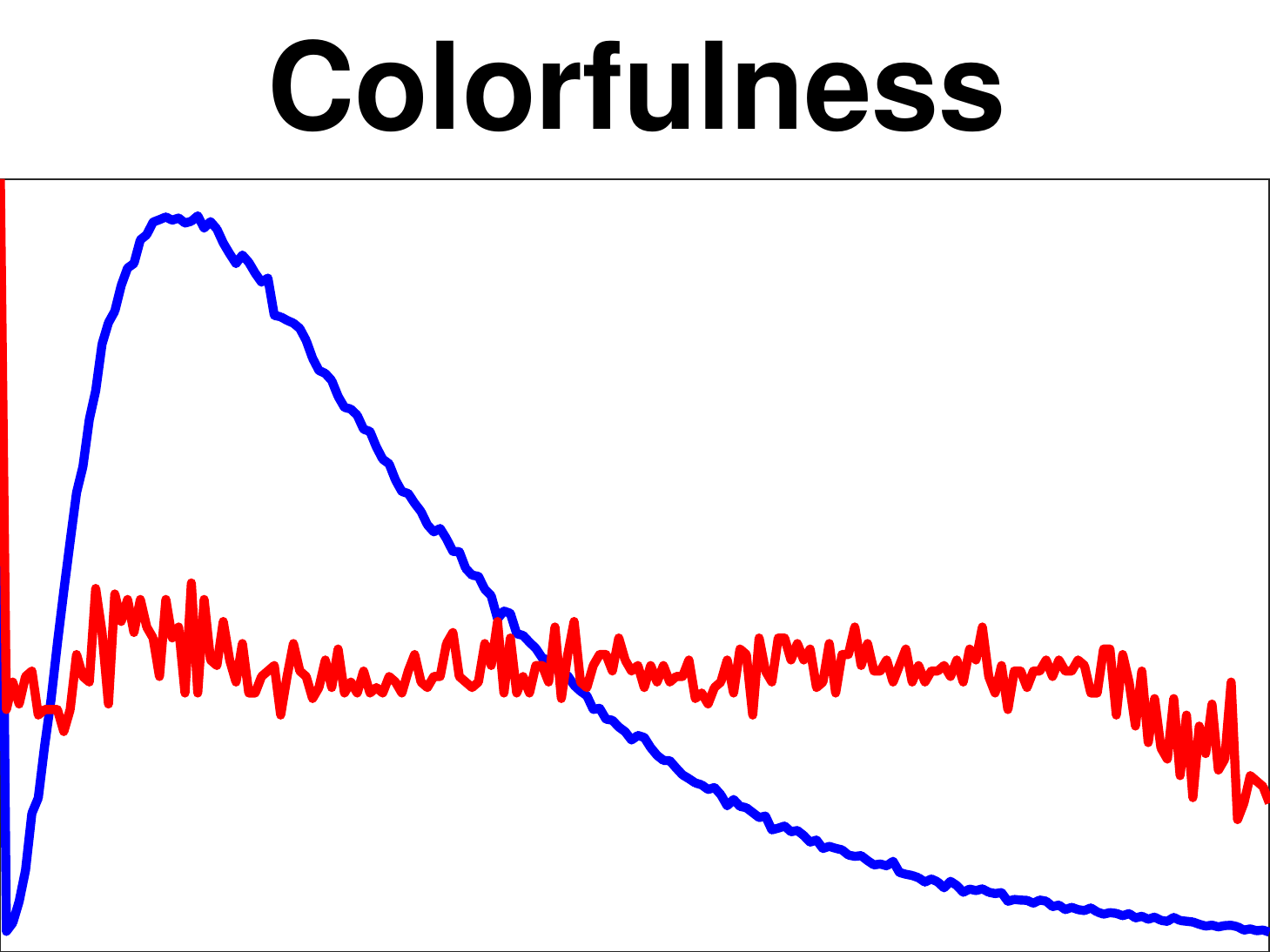}}
\end{minipage}
\begin{minipage}{0.119\linewidth}
\centerline{\includegraphics[width=\textwidth]{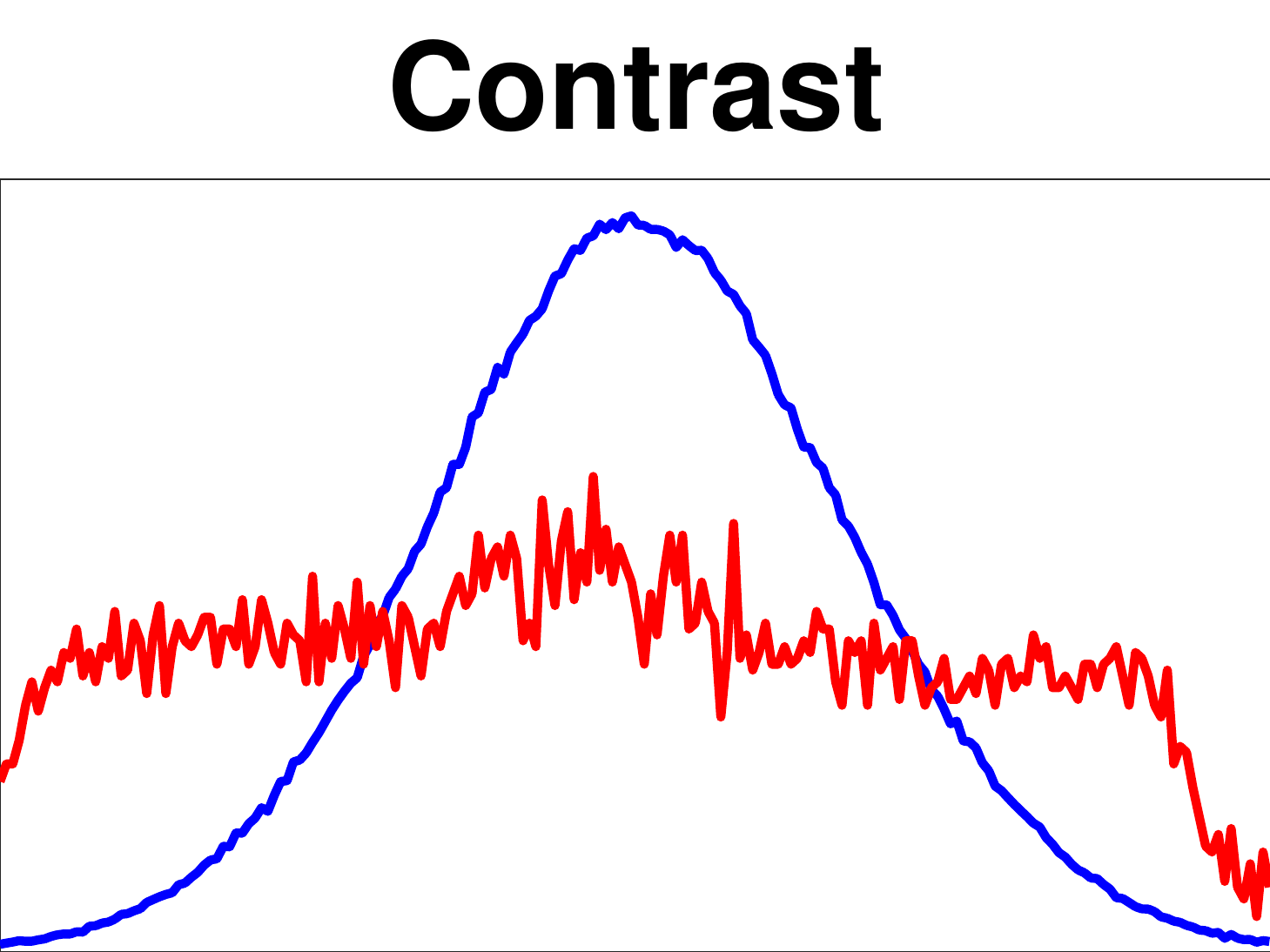}}
\end{minipage}
\begin{minipage}{0.119\linewidth}
\centerline{\includegraphics[width=\textwidth]{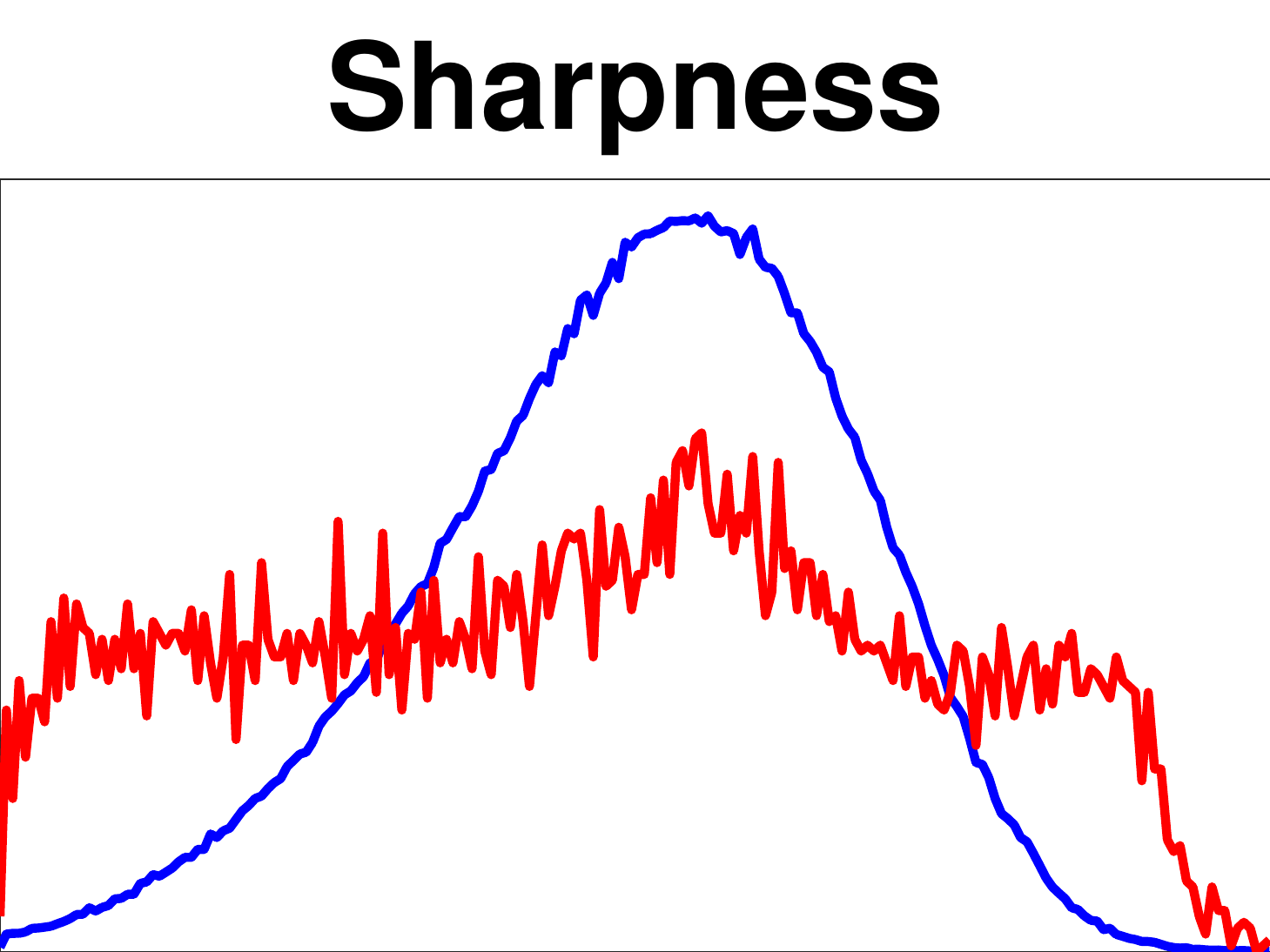}}
\end{minipage}
\begin{minipage}{0.119\linewidth}
\centerline{\includegraphics[width=\textwidth]{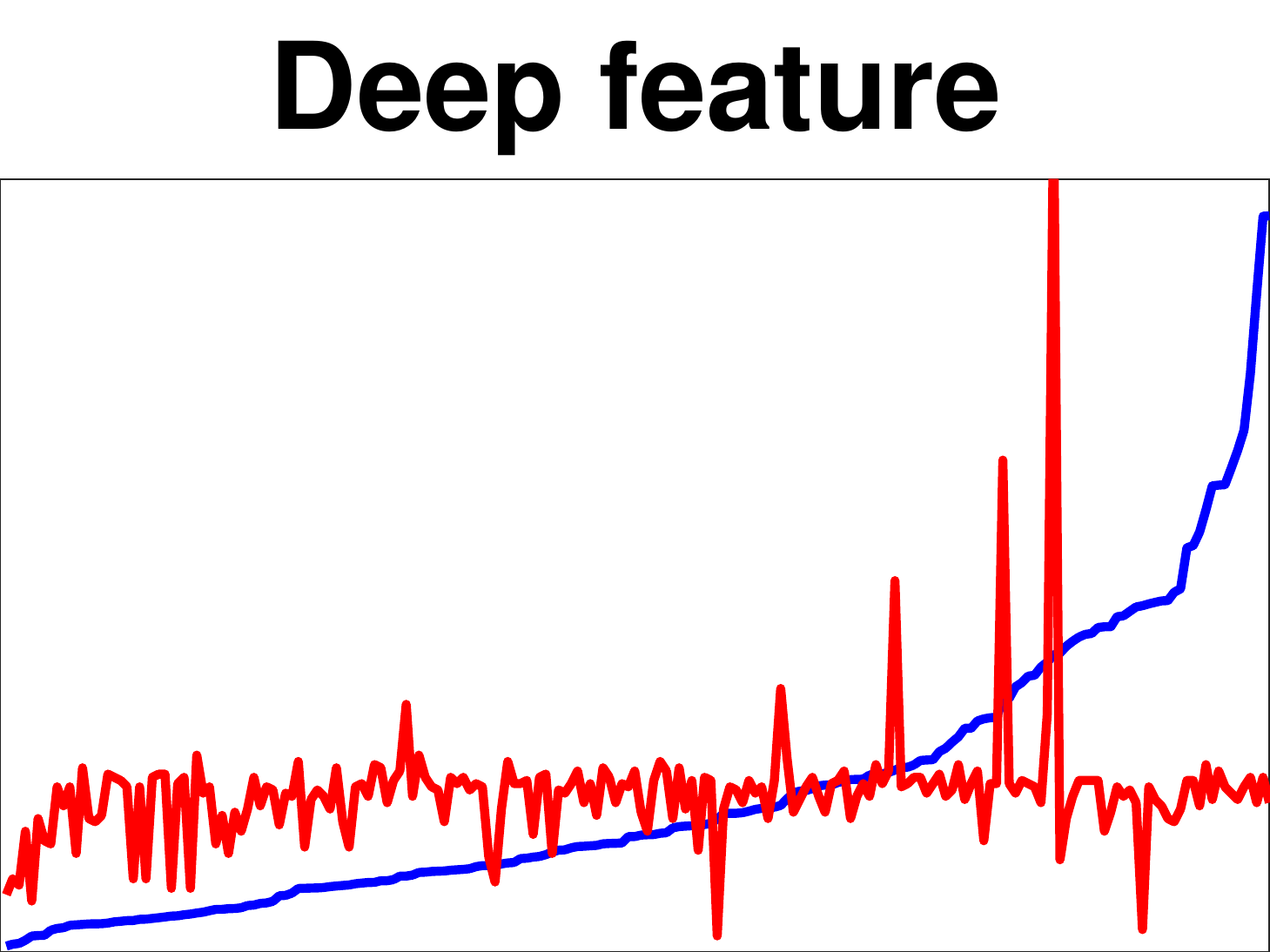}}
\end{minipage}
\caption{The distribution of indicator values as extracted from the larger subset of 866,976 YFCC100m images (blue) and from the sampled 10,073 images (red). The sampling procedure enforces a more uniform distribution on each indicator after post-processing compared to the original distribution.}
%
%
%
%
\label{fig:unisample}
\end{figure*}

\subsection{Selective image cropping}
\label{sec:selective_cropping}
It is important to standardize the resolution of all images in our database. It applies to user studies, computing various measures, and using the images for training, or benchmarking IQA methods. We chose $1024\times768$ pixels as a standard resolution, as we found that 95\% of our crowd workers' devices have at least this resolution. Rather than shrinking images unevenly and thus changing their aspect ratio, we cropped them, as changing an image's aspect ratio affects its perceived quality and reduces the authenticity of the image collection.

A naive approach to cropping images is choosing their central region and removing the rest. After trying this strategy in an early experiment, we observed that wrong crops such as those that remove essential parts of the image, like faces, reduce the perceived quality of images compared to the case where the entire image is presented. Hence, we devised a selective approach that does not create some of the more frequent unintended degradations.

\begin{figure}[!htbp]
\centering
\includegraphics[width=0.5\textwidth]{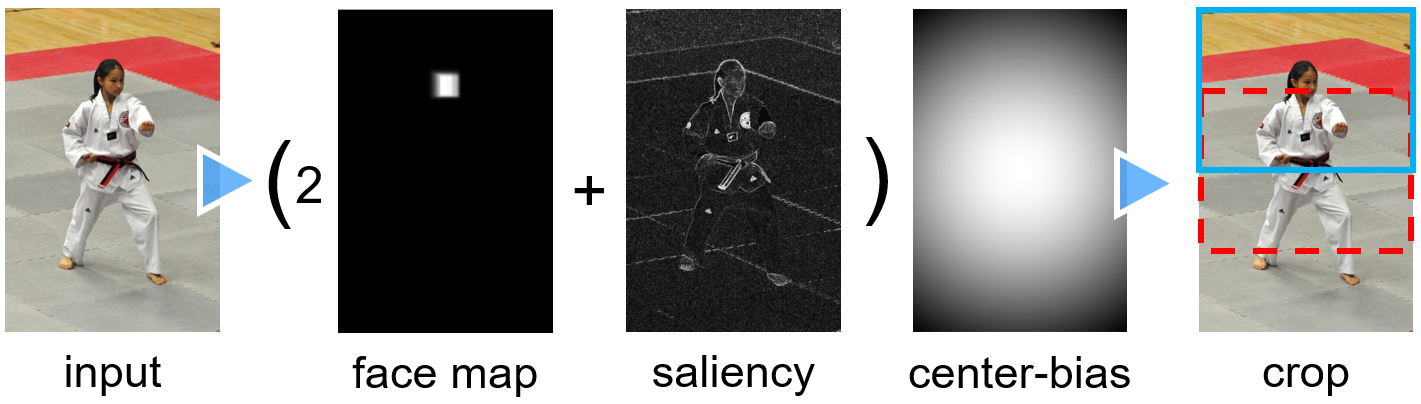}
\caption{We cropped each image in our database to a standard size of $1024\times768$ by accounting for the presence of faces, saliency, and a center-bias. The combination of the three middle maps forms the importance map. The cropping result, shown in light blue (solid line) on the rightmost image correctly includes the person's face, which would have been otherwise removed from the picture using a naive centered crop (dashed, red line).}
\label{fig:smart_crop}
\vspace{-10pt}
\end{figure}

The aim was to keep faces and salient areas inside the crop and to process one million images sufficiently fast. We combined the Viola-Jones face detector \cite{viola2004robust} over multiple poses (front, profile left, and right) with a saliency detection method \cite{hou_image_2012} into an importance map, see Fig.~\ref{fig:smart_crop}. The authors in \cite{hou_image_2012} argued that experimental data shows their method identifies foreground content well, which is important for us.  Sharper areas are more likely to be relevant for image quality -- for instance, in a case where an image has out of focus regions, and we would rather focus on the subject. Last, but not least, the saliency method we used has a small computational cost. 

The crop had to maximize the mean importance. We chose the crop location by convolving a kernel the size of $1024\times768$ with the importance map. The kernel value was $1$ everywhere except for a border of  $10$ pixels on all sides where it was $-1$. The convolution was efficiently done in the frequency domain leading to an average crop time below one second per image on a standard desktop CPU (2.4 GHz Xeon).

\subsection{Diversity sampling}

Our objective was to select a subset of images while ensuring the diversity of content and distortion authenticity. The latter is implicit due to the source of the images. We ensured content diversity using a sampling procedure that relies on quality-related indicators and a category-features. 


\subsubsection{Image quality indicator selection}

We collected several image quality indicators relating to brightness, colorfulness, contrast, noise, sharpness, and No-Reference (NR) IQA measures. For some of them, several implementations were available. We discarded those that were too slow to be run on our set of one million images. We conducted preliminary subjective studies and kept four measures that are well-correlated with human perception, namely brightness, colorfulness \cite{Hasler:2003}, Root Mean Square (RMS) contrast, and sharpness \cite{Vu:2012}. Besides these, we considered three other indicators: image bitrate, resolution (height$\times$width), and JPEG compression quality (image meta-data); these are highly correlated with image quality.

Each quality indicator identifies an image attribute, measuring its magnitude or presence as a scalar value. Extreme values for an indicator relate to severe distortion, either due to the absence or abnormal emphasis on that particular aspect. If we were to sample our image database randomly, it is unlikely that images having ``abnormal" attribute values would be selected. Therefore, we sampled images with a broader range of indicator values, and thus potentially more distortions types. 

Nonetheless, the absolute extremes of the indicator ranges are distorted to an excessive degree and were uninformative, i.e., overly dark or bright, overly colorful, and others. Therefore, before performing the sampling procedure, we trimmed the extreme ends of each indicator distribution. We computed the z-score of an indicator value $x$ as $z = (x-\bar{X})/{S}$, where $\bar{X}$ and $S$ are the mean and standard deviation of a sample for the given indicator. By removing all images with an absolute z-scored indicator value greater than 3, the dataset size shrank from one million to 866,976.

\subsubsection{Choice of content indicator}

Until this point, we ensured content diversity by sampling one million images based on machine tags provided by YFCC100m. The tags had been assigned using a deep neural network for classification and represented a few most likely categories per image. Furthermore, we selected a subset of 866,976 images, such that to exclude extreme quality indicator values.

To further improve the content description, we made use of more comprehensive $4096$-dimensional deep-features extracted from a VGG-16 model pre-trained on ImageNet \cite{vgg}. The features represent the activations of the last fully-connected layer (usually named FC7) before the final layer that returns class likelihoods.

Since the deep-features are 4096-dimensional vectors, we applied a bag-of-words model to quantize them. That is, we ran $k$-means to compute 200 centroids, mapping each deep-feature to the nearest cluster. The cluster indices came to represent our content indicator.

\subsubsection{Sampling strategy}

For the actual sampling, we applied the method proposed by Vonikakis et al.~\cite{vonikakis2016shaping}, enforcing a uniform target distribution for each indicator. For the quality indicators, we quantize each indicator value into $N$ bins. The content indicator is already quantized as the cluster index. The sampling procedure jointly optimizes the shape of the histograms along all indicator dimensions, using Mixed Integer Linear Programming (MILP).

We used $N=200$ bins for all seven scalar indicators. We ran the sampling procedure -- generating 13,000 images -- with uniformly sampled indicators. The set is larger than the target of 10,000 to allow for removing duplicates and other post-filtering.

\subsection{Removal of duplicates and inappropriate content}

The uniform sampling strategy ensures the diversity of the image database on a broad scale. However, due to the binning procedure, identical copies or near-duplicate images can be sampled together, i.e.,  photos of a scene taken from slightly different views. 


We devised a way to remove near-duplicates. First, the values of each indicator were scaled to the interval $[0,1]$. We computed all pairwise euclidean distances $D(i,j)$ between images $i,j$ from the source dataset in the 8-dimensional indicator, plus content space. The distance in the content space is set to 0 if two images are part of the same cluster, and one otherwise. Duplicate and near-duplicate images $i,j$ are expected to correspond to small distances $D(i,j)$. Thus, by iteratively removing a member of the closest pair, we can effectively remove near-duplicates. We removed 2,000 images in this way.\footnote{See Fig.~14, supplementary material, for examples.}

To ensure the quality of our database, we manually removed images showing too little content, namely text screenshots, text scans, heavily under-exposed images, or inappropriate images showing mature content.\footnote{See Fig.~15, supplementary material, for examples.}
In the end, 10,073 images remained to make up our KonIQ-10k database; see Fig.~\ref{fig:unisample}.

\begin{figure*}[t]
\centering
\begin{minipage}{0.2\linewidth}
\centerline{\includegraphics[width=\textwidth,height=70pt]{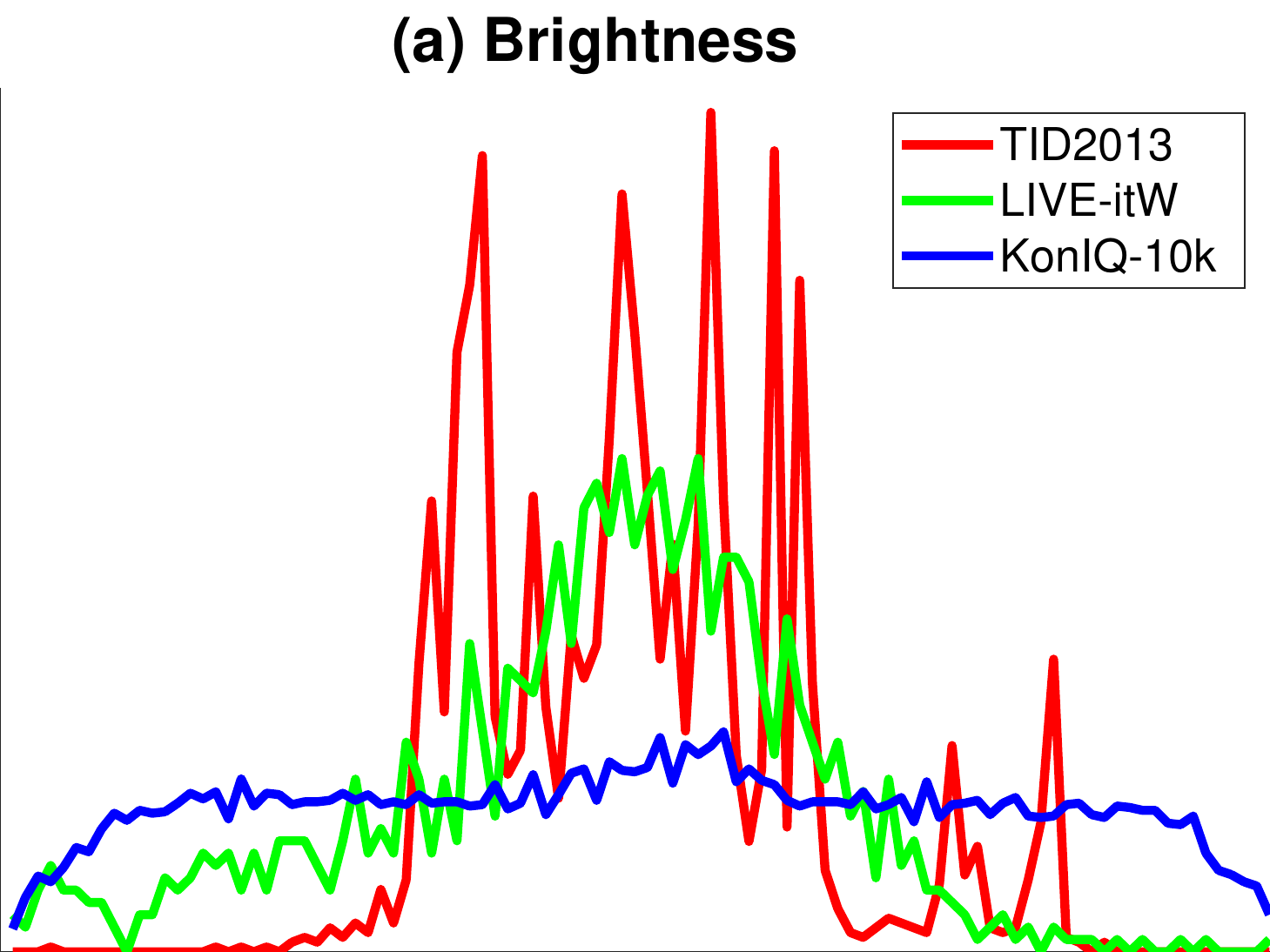}}
\end{minipage}
\hspace{8pt}
\begin{minipage}{0.2\linewidth}
\centerline{\includegraphics[width=\textwidth,height=70pt]{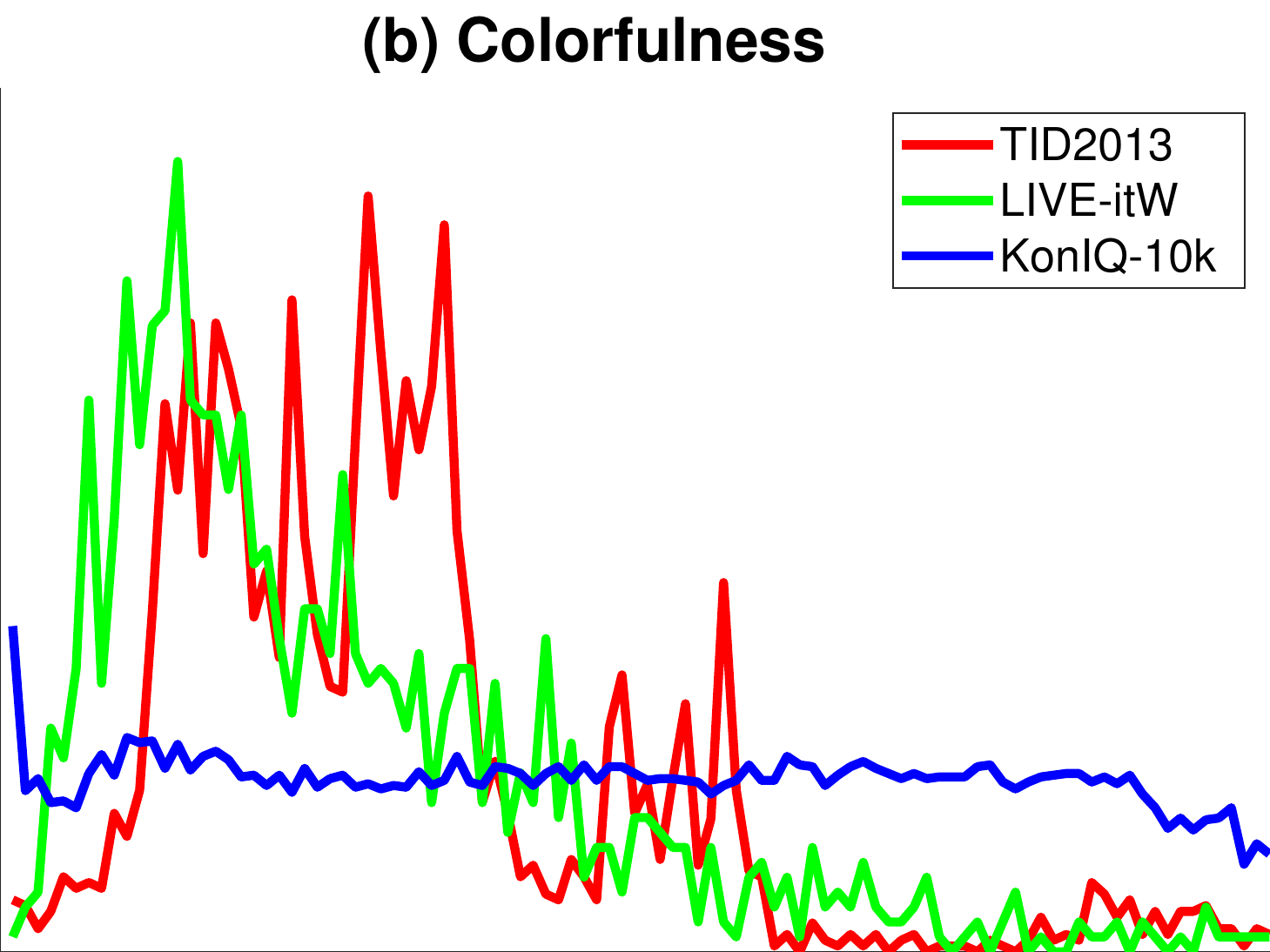}}
\end{minipage}
\hspace{8pt}
\begin{minipage}{0.2\linewidth}
\centerline{\includegraphics[width=\textwidth,height=70pt]{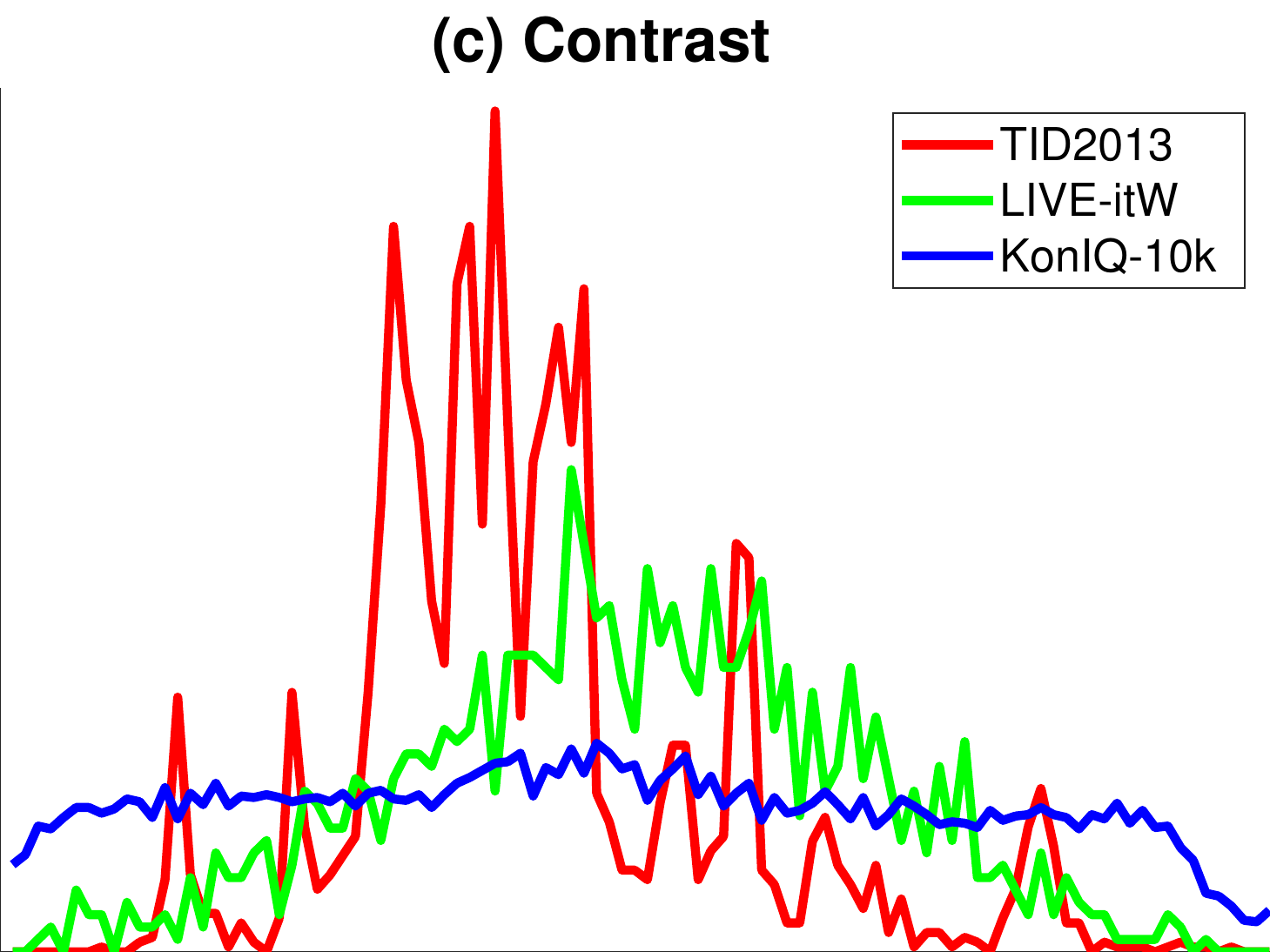}}
\end{minipage}
\hspace{8pt}
\begin{minipage}{0.2\linewidth}
\centerline{\includegraphics[width=\textwidth,height=70pt]{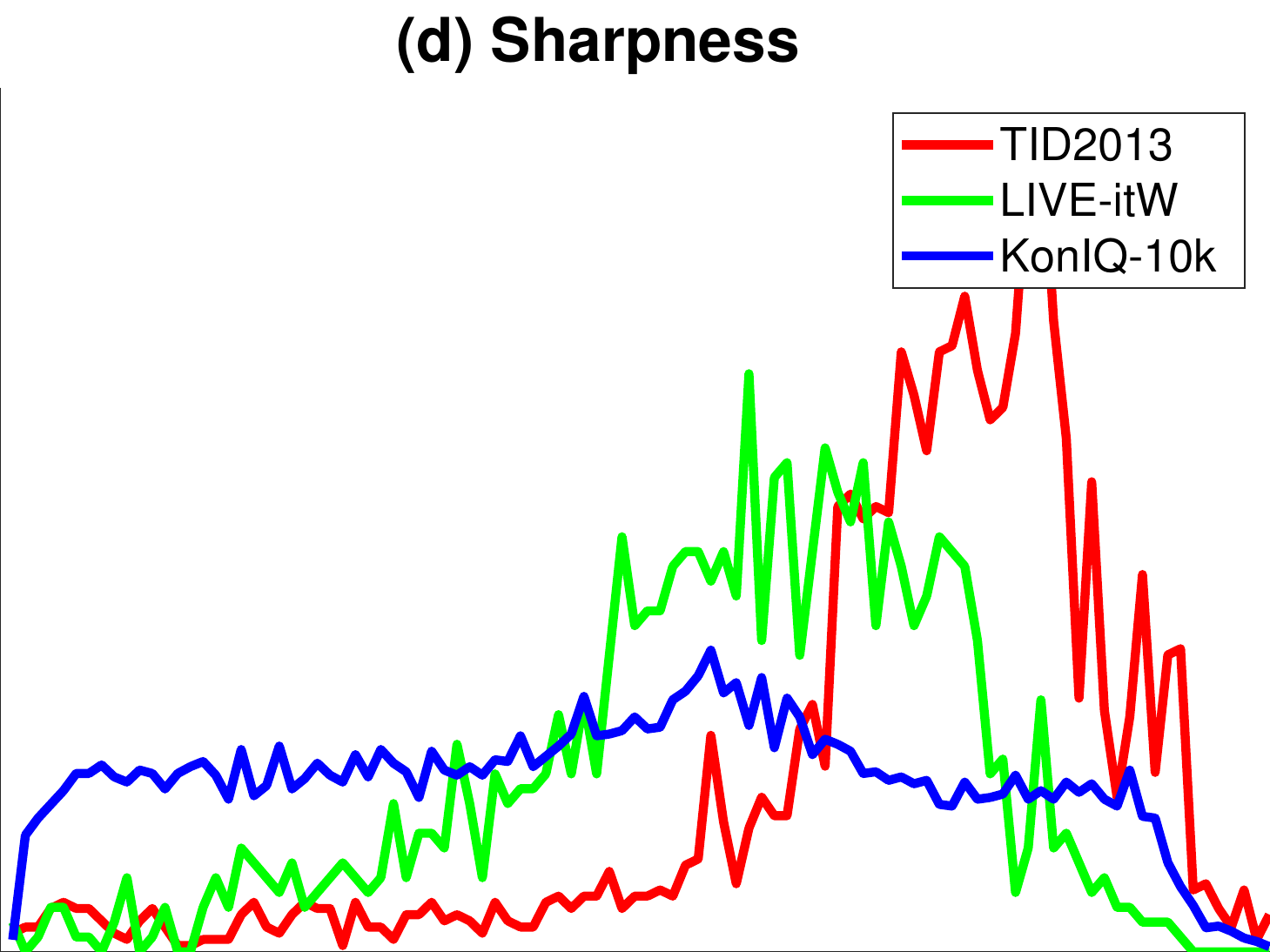}}
\end{minipage}\\
\vspace{8pt}
\begin{minipage}{0.2\linewidth}
\centerline{\includegraphics[width=\textwidth,height=70pt]{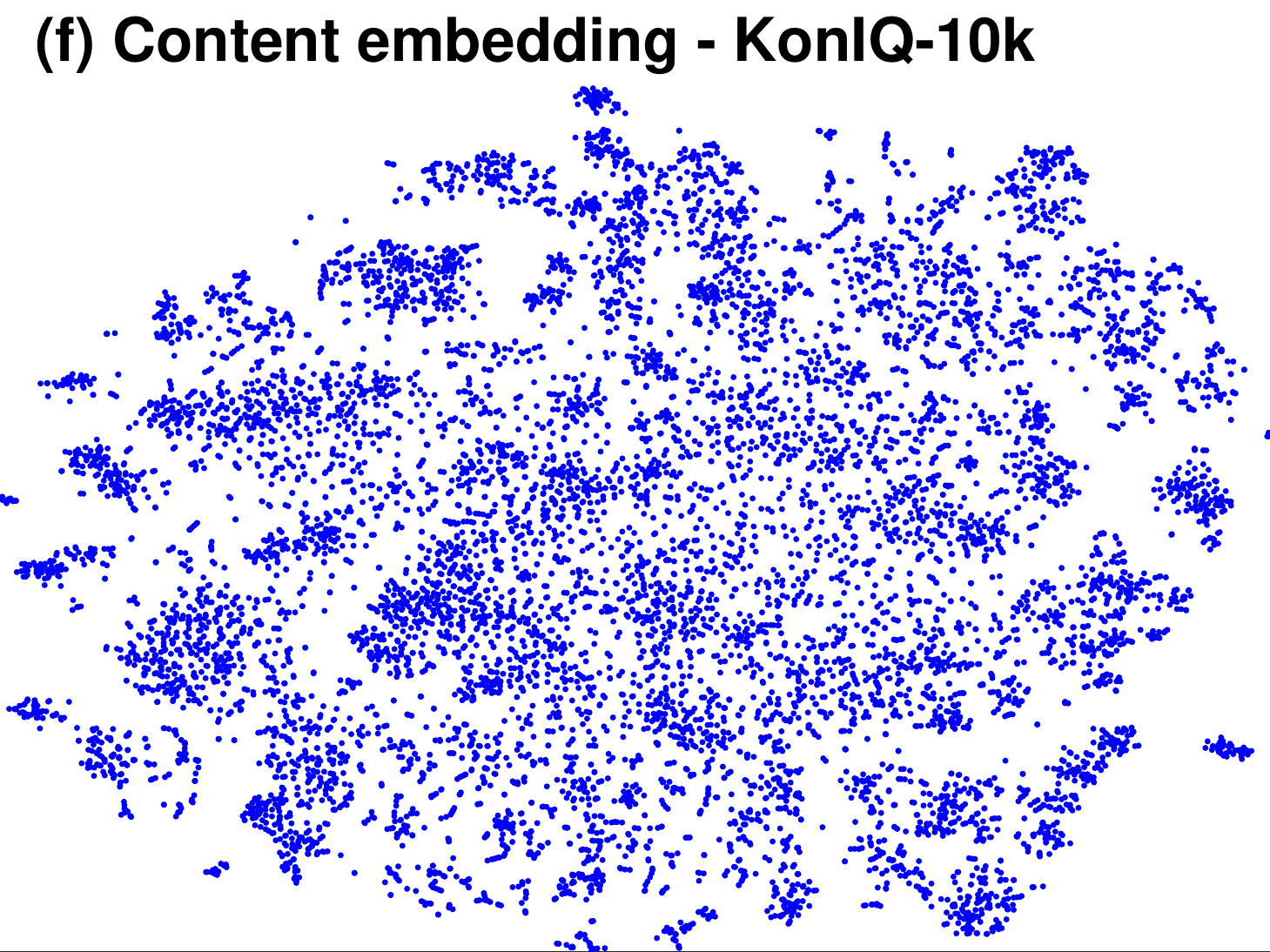}}
\end{minipage}
\hspace{8pt}
\begin{minipage}{0.2\linewidth}
\centerline{\includegraphics[width=\textwidth,height=70pt]{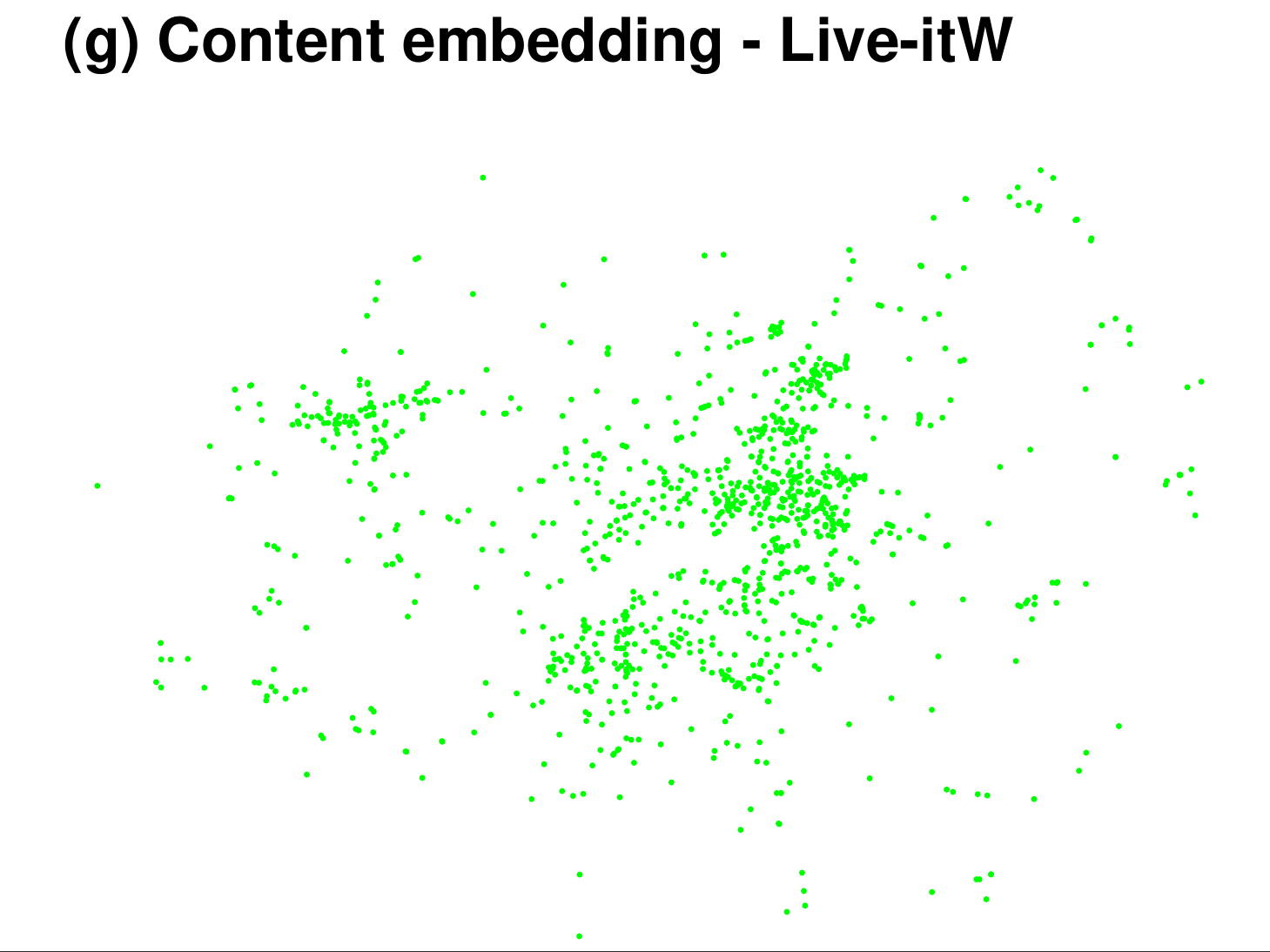}}
\end{minipage}
\hspace{8pt}
\begin{minipage}{0.2\linewidth}
\centerline{\includegraphics[width=\textwidth,height=70pt]{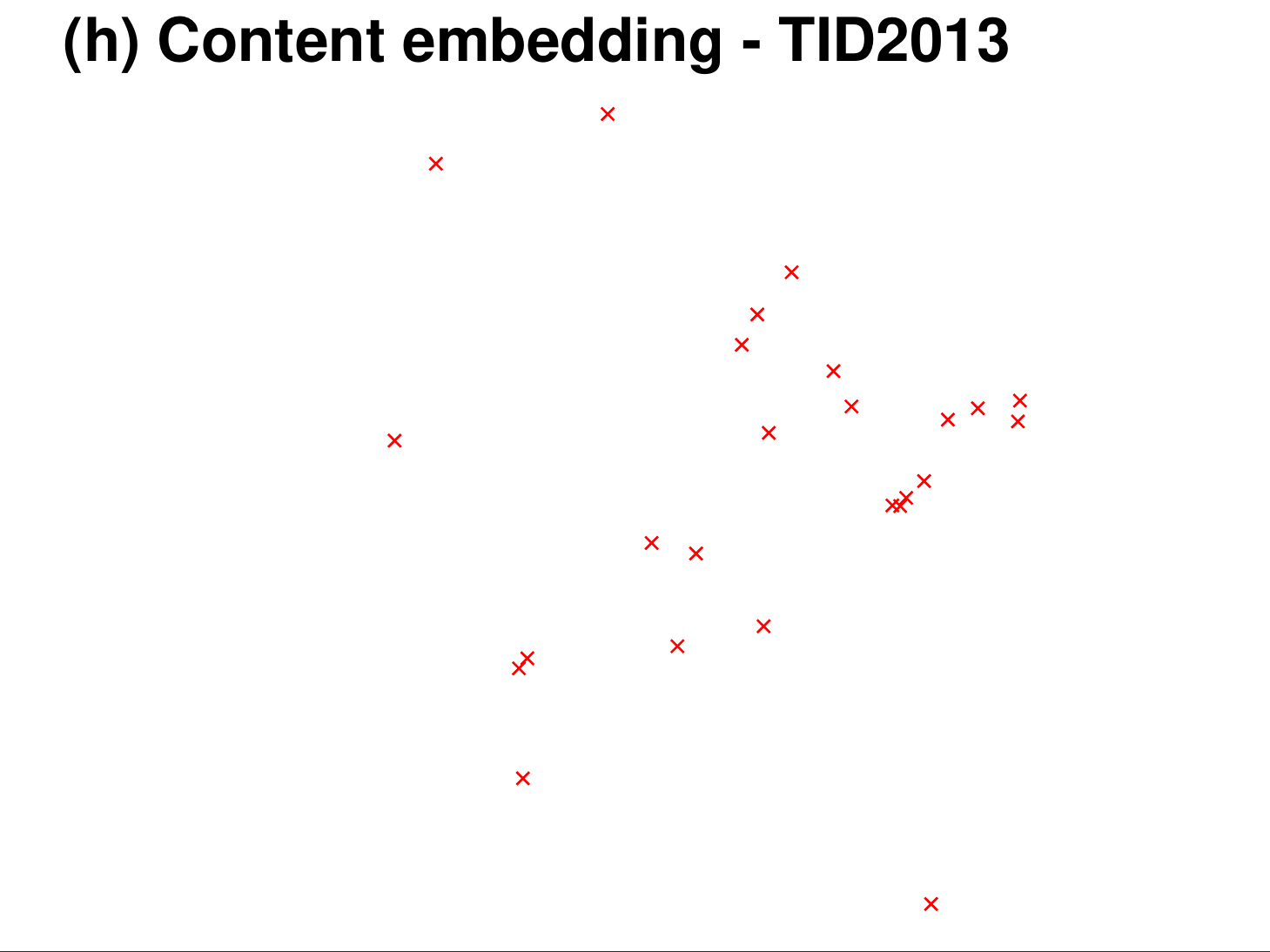}}
\end{minipage}
\hspace{8pt}
\begin{minipage}{0.2\linewidth}
\centerline{\includegraphics[width=\textwidth,height=70pt]{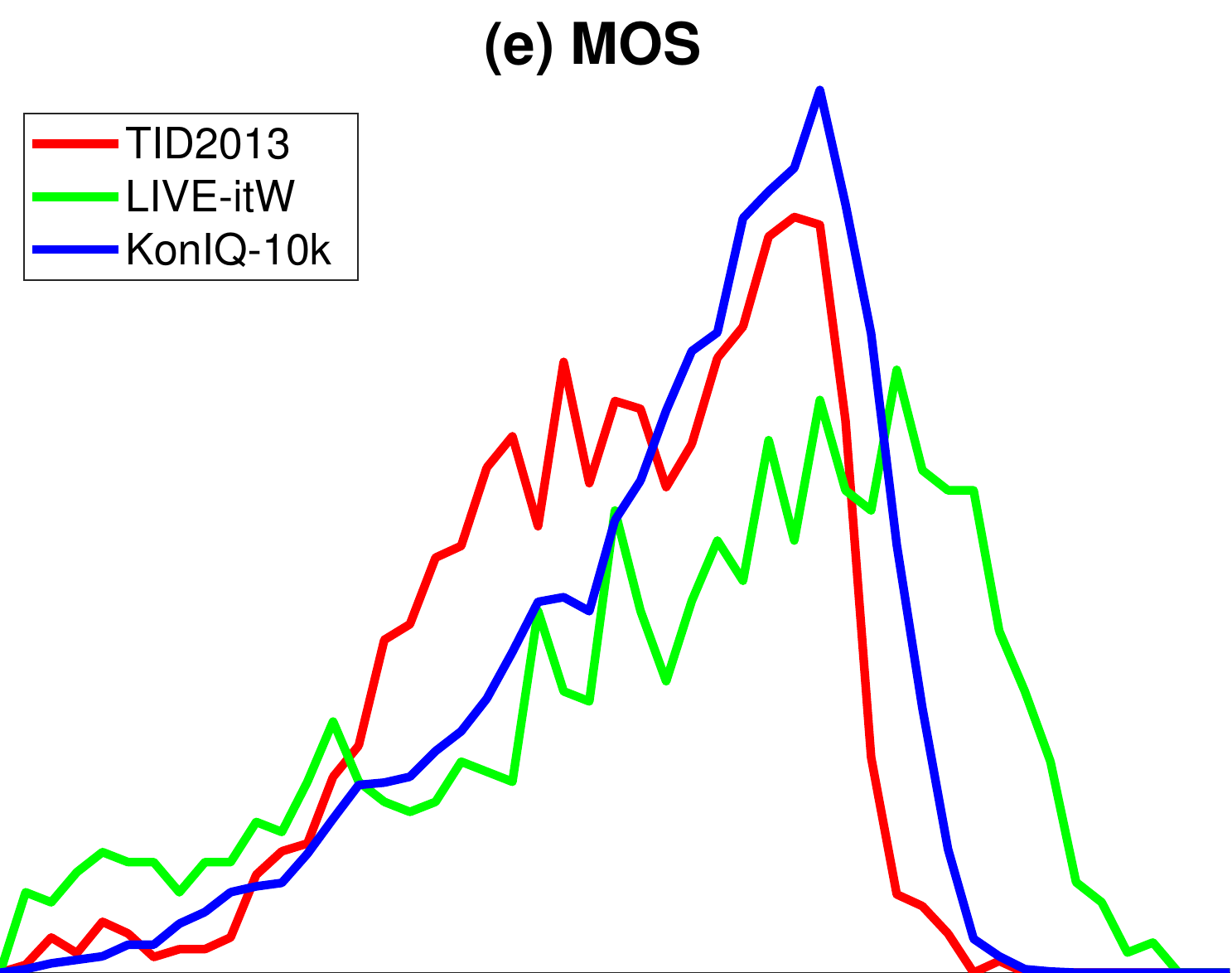}}
\end{minipage}
\caption{Diversity comparison between TID2013, LIVE-itW, and KonIQ-10k. (a) - (d) distribution comparison in brightness, colorfulness, contrast, and sharpness, respectively. (f) - (h) deep feature embedding in 2D via t-SNE. (e) MOS distribution.}
\label{fig:diversitycomp}
\vspace{-10pt}
\end{figure*}

\subsection{Diversity analysis}


We selected LIVE-itW and TID2013 to compare their diversity with KonIQ-10k in some aspects. Here, LIVE-itW and TID2013 are the most representative, authentically distorted, and artificially distorted databases. Their distributions in brightness, colorfulness, contrast, and sharpness are depicted in Fig.~\ref{fig:diversitycomp}(a)-(d), respectively. KonIQ-10k is more diverse than compared databases in each of those indicators. Examples of images of KonIQ-10k are shown in Fig.~17 (Supplementary Material), where five images are uniformly sampled in each of the four indicators. 

To compare the content diversity, we embedded the 4096-dimensional VGG-16 deep features from the databases into a 2D subspace by t-SNE \cite{van2008visualizing}. The visualization is shown in Fig.~\ref{fig:diversitycomp}(f)-(h). Since just a few photographers captured LIVE-itW images, their content only covered a small region of KonIQ-10k, not to mention TID 2013, which was generated from only 25 reference images. After aligning the scales, their MOS distributions are illustrated in Fig.~\ref{fig:diversitycomp}(e). Examples of images with uniformly sampled MOS are shown in Fig.~18 (Supplementary Material).

\section{Subjective Image Quality Assessment} 


To assess the visual quality of the 10,073 selected images, we performed a large-scale crowdsourcing experiment at CrowdFlower.com (now Figure-Eight.com). The experiment first presented workers with a set of instructions, including the definition of ``technical image quality", some considerations when giving ratings, examples of often encountered distortion types, and images with different qualities. 
The subjects were then instructed to consider the following types of degradations: noise, JPEG artifacts, aliasing, lens and motion blur, over-sharpening, wrong exposure, color fringing, and over-saturation. We used the standard 5-point Absolute Category Rating (ACR) scale, i.e., bad (1), poor (2), fair (3), good (4), and excellent~(5). 


\subsection{Domain experts and test questions}
We devised a set of test questions, i.e., rating questions with known answers, to filter reliable crowd workers and for analyzing the overall quality of the collected ratings. The opinions of domain experts are generally more reliable, and thus provide a good source of information for setting test questions. We involved 11 freelance photographers who had, on average, more than three years of professional experience. We asked them to rate the quality of 240 images: 29 were pristine high-quality images, carefully selected beforehand, 21 were artificially degraded using 12 types of distortions, and the remaining 190 images were randomly selected from Flickr (not part of our 10k dataset). The distortions included blur, artifacts, contrast, and color degradation. 

Based on this set of images and the mean opinion score from the freelancers, we generated test questions for our crowdsourcing experiment. The correct answers were based on the rounded values of the freelancers' MOS $\pm$ one standard deviation. We allowed a margin for error, tolerating some subjectivity in the participants' rating behavior.  As a result, all images had, at most, three valid answer choices.

\subsection{Crowdsourcing experiment}

A total of 2,302 crowd workers participated in the study, which took three weeks to complete.  Several filtering steps were implemented and validated to ensure an acceptable level of quality for the resulting mean opinion scores (MOS).

{\smallskip \noindent \textbf{Quiz}.} Before starting the actual experiment, workers took a quiz consisting of 20 test questions. Only 1,749 workers with an accuracy over 70\% were eligible to continue.  

{\smallskip \noindent \textbf{Hidden test questions}.} Hidden test questions were presented throughout the main part of the experiment to encourage contributors always to pay full attention. Again, only workers with an accuracy above 70\% were allowed to complete their jobs, leaving 1,648 remaining contributors. Their ratings yielded a preliminary set of MOS values. 

{\smallskip \noindent \textbf{Outliers}.} Workers who had a very low agreement with the preliminary MOS were regarded as outliers. 68 workers with a PLCC of their votes and the preliminary MOS lower than $0.5$ were removed from the study.

{\smallskip \noindent \textbf{Line clickers}.} Line clickers are workers with an unusually high frequency for any single answer choice, in the context of a multiple-choice questionnaire \cite{QoMEXReliability}. To detect line clickers, we computed the score counts of each worker for all five answer choices. We then took the ratio between the maximum count and the sum of the four lower counts. 121 workers with a ratio larger than $2.0$ were removed.

\smallskip
All in all, we arrived at 1,459 of 2,302 crowd workers who passed these filtering steps. As a result, to annotate the entire database of 10,073 images, with at least 120 scores each, more than 1.2 million trusted judgments were submitted. 

Finally, to compensate for differing ranges of the ratings of individual workers, we rescaled their scores to $[1,100]$ by
$$
s_{ij}^{\text{norm}} = 1 + 99 \frac {s_{ij} - \min_j (s_{ij})}{\max_j (s_{ij}) - \min_j (s_{ij})},
$$
where $s_{ij}$ is the score of the $i$-th worker on the $j$-th image. 

To check the reliability of the crowd MOS, we compared them to those obtained from the group of 11 experts. Out of all 240 images for the test questions, we have 187 images, which had each been rated by 11 experts and 592 or more crowd workers. We can regard the expert opinion ($\text{MOS}_\text{experts}$) as ``ground truth'' and accept the crowdsourcing results as reliable when for the vast majority of these 187 images, the crowd MOS values are enclosed in the 95\% confidence interval of the experts' MOS.

We compensated for the difference in the range of the MOS between the two data sources by fitting a linear model: 
$$\text{MOS}_\text{experts} = 1.12\cdot\text{MOS}_\text{crowd} - 10.43.$$ Note that the alignment of the crowd to the expert MOS scale is not applied to transform the final scores that are part of the database.

After the re-alignment, we compared the two data sources: experts and crowd. For the crowd data, we bootstrapped MOS computations for subsamples of size of up to 120 -- the number of ratings in KonIQ-10k. Each set of MOS values was compared to the ground truth MOS, giving rise to a root-mean-square error (RMSE), see Fig.~\ref{fig:rmse_curves}(a). We found that this RMSE quickly converges to a lower bound of 11.35 on the 100 point scale. We also bootstrapped expert groups of size 11 and found a standard deviation of bootstrapped MOS values of 6.63.





\subsection{Reliability of the crowd}
\label{sec:reliability}

%
%
%
%
%

\begin{figure}[!t]
\centering
\begin{minipage}{.46\linewidth}
\centerline{(a) average errors}
\centerline{\includegraphics[width=\textwidth]{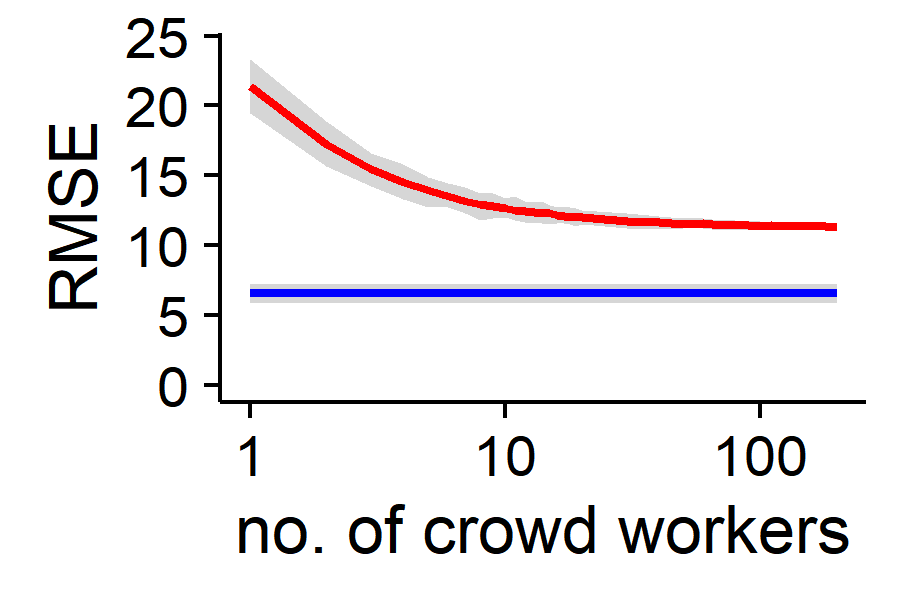}}
\end{minipage}%
\begin{minipage}{.46\linewidth}
\centerline{(b) per image errors}
\centerline{\includegraphics[width=\textwidth]{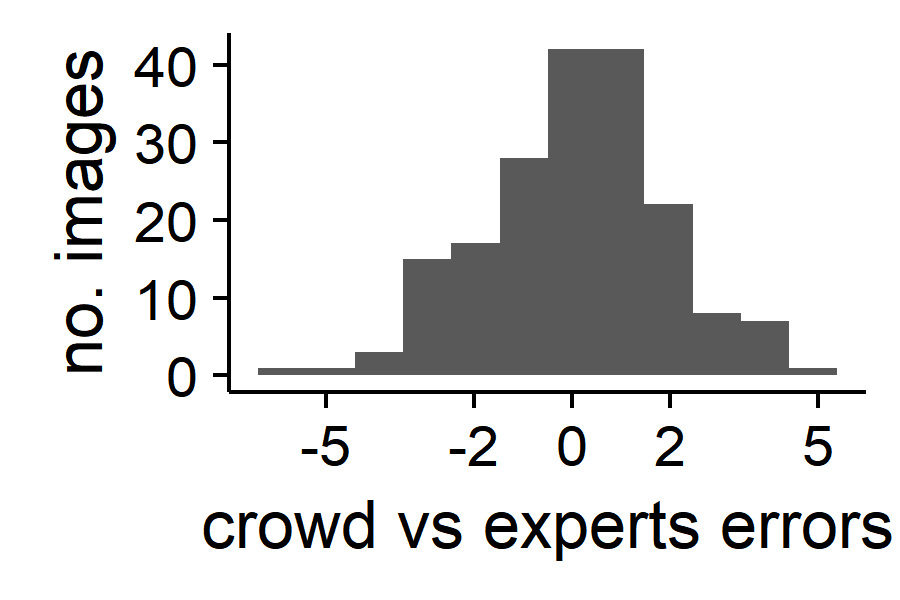}}
\end{minipage}

\caption{(a) Top red line: bootstrapped RMSE of crowd  MOS against MOS of 11 experts; Bottom blue line: bootstrapped standard deviation of MOS of 11 experts; gray ribbon is the 95\% CI of the RMSE. (b) Distributions of errors of crowd MOS against experts' MOS, expressed in multiples of the standard deviation of the bootstrapped MOS of 11 experts.}
\label{fig:rmse_curves}
\end{figure}
%
%



In Fig.~\ref{fig:rmse_curves}(b), we show the distribution of the errors of the crowd MOS over all 187 images. We note that for 137 out of 187 images (73\%), the errors are within the $\pm 2$ standard deviations of the experts' MOS (95\% confidence interval). Therefore, about three-quarters of the images are sufficiently well-rated by the crowd so that they can be confused with the ratings of experts.

The crowd MOS on the remaining 50 images diverges more from the experts. A preliminary inspection shows that 11 of the 27 items that the crowd rated lower than the experts represent shallower depth-of-field images, e.g., the first five in Fig.~16 (Supplementary Material). 
Generally, for crowd workers, a large amount of blur was considered a significant degradation, whereas the professional photographers understood it as an artistic effect, which did not reduce the quality. The observed disagreement is, at least in part, a consequence of diverging domain knowledge between the expert (freelancers) and novice (crowd) groups.


Furthermore, in support of the reliability of our experiment, the work of Hosu et al.\ \cite{QoMEXReliability} has shown that screening users based on image-quality-test questions improves the intra-class correlation coefficient (ICC). They have found that their IQA experiment ICC increased from 0.37 to 0.50 by requiring 70\% accuracy on quality-based test questions. The approach in our work here has a similar effect, leading to an ICC of 0.46 on the entire database. This suggests KonIQ-10k has a better result than similar crowdsourcing assessment studies, such as those of Redi et al.~\cite{siahaan_reliable_2016}. In the latter work, the authors performed an aesthetic quality assessment for abstract images. After careful reliability checking, it achieved an ICC of 0.403 for the ACR scale.
%
%
%
%
%

\begin{figure}[!t]
\centering
\includegraphics[width=0.8\linewidth]{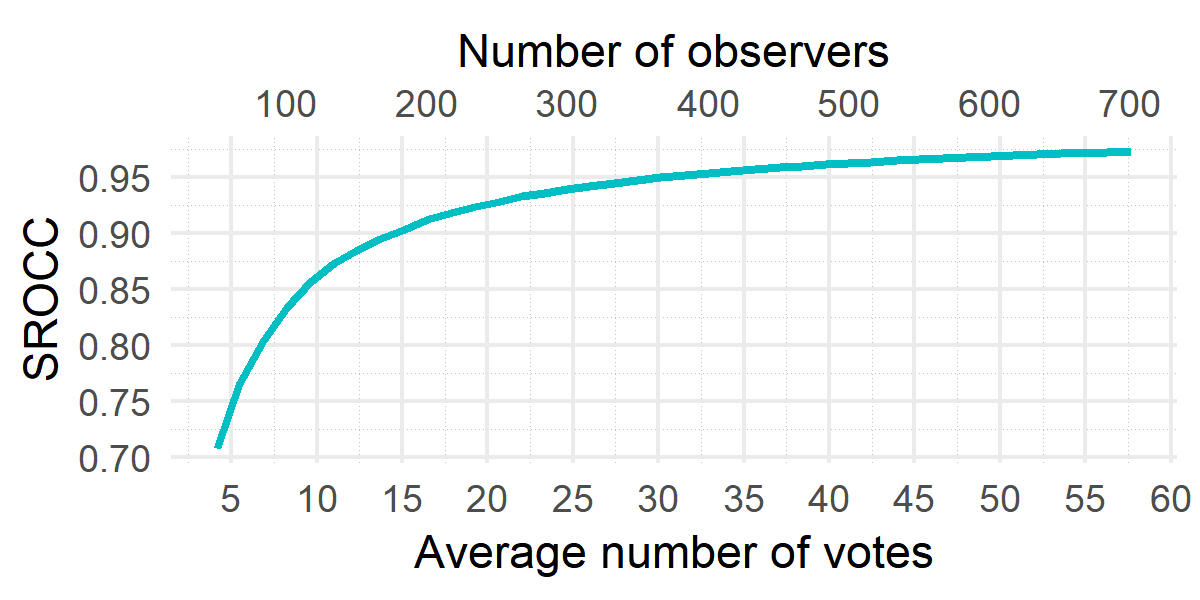}
\caption{Mean agreement (SROCC) between MOS values when the number of observers increases, and thus the average number of votes per image. 
The agreement between group sizes of 700 observers reaches $0.973 \pm 0.001$ SROCC. In this case, the average number of ratings per image is $57.68$.}
\label{fig:group-agreement}
\end{figure}

Another relevant metric for the consistency of the scoring produce is the mean inter-group agreement. In Fig.\ \ref{fig:group-agreement}, we show that the mean agreement between the MOS values of non-overlapping random groups of users increases as the number of users grows. When considering the correlation between random halves of the contributors in our experiment, the mean SROCC reaches an excellent value of 0.973.

For further analysis, we will provide the complete raw crowdsourcing data as part of the final release of our database. This includes information about observer context, such as screen resolution, browser zoom, answer timings, and more.

\section{Finding a better end-to-end deep BIQA architecture}

Existing BIQA approaches rely on standard CNN architectures designed for image classification (ImageNet \cite{deng2009imagenet}). Some of the main factors that have been considered in the design of better methods are 1. The way the input images are presented to the network, such as down-sized versions of the original image or crops; 2. The choices for the base architecture; 3. The loss function to be minimized; 4. The aggregation strategy, in case multiple predictions are made, such as from multiple crops of the same image. Even though several existing works have shown promising results, there is no definite answer to which particular combination of factors is most suitable for BIQA when training end-to-end.

Recent works, such as DeepRN \cite{varga2018deeprn}, showed that training on large-resolution images, rather than downsized or cropped versions, improved prediction performance. Moreover, training to predict distributions of ratings compared to only MOS (a scalar) had been suggested to work better for BIQA \cite{talebi_nima_2017,varga2018deeprn}. Various loss functions had been considered as well, including regression on scalars, such as MAE and MSE, but also distributional losses, Earth Mover's Distance, and Huber loss on distributions being an example. Thus, we studied the effect of input size, loss function, and their combinations. In addition to the frequently used base architectures in previous works, such as VGG16 \cite{vgg}, InceptionV3 \cite{szegedy2017inception}, and ResNet101 \cite{he2016deep}, we considered the performance of the more modern InceptionResNetV2 \cite{szegedy2016rethinking} and NasNetMobile \cite{zoph2017learning}. Our study aims to show how close we have gotten to solving BIQA in the wild. Moreover, we tested the performance both on KonIQ-10k (test set) and cross-tested on LIVE-itW to get a better understanding of the generalization potential of each approach. 

%
%

\subsection{The proposed architecture}

\begin{figure*}[!t]
\centering
\includegraphics[width=0.7\linewidth]{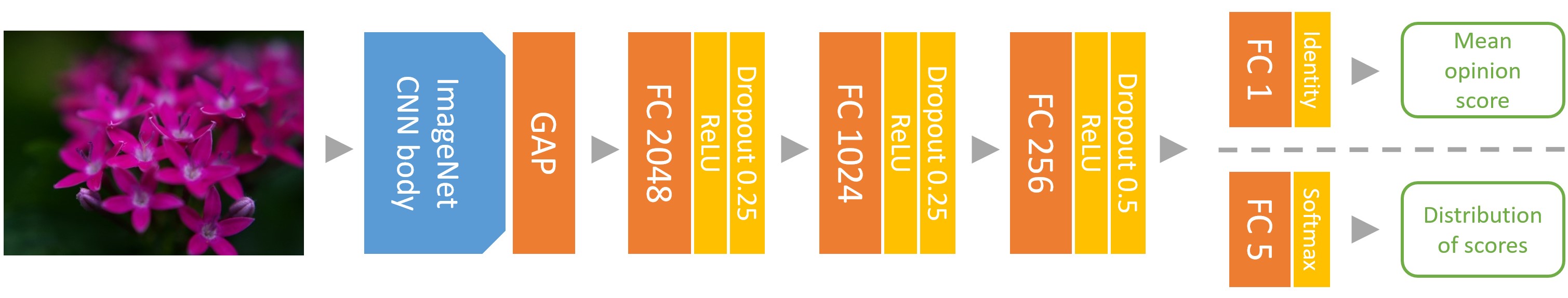}
\caption{Architecture of the end-to-end system. We test two types of training, one predicting the MOS values and the other predicting the distribution of ratings.}
\label{fig:architecture-end-to-end}
\vspace{-10pt}
\end{figure*}


The architecture of our proposed end-to-end method is displayed in Fig.~\ref{fig:architecture-end-to-end}. Given an image, it is passed through a state-of-the-art CNN body (the convolutional layers, without the final fully-connected layers), followed by a Global Average Pooling (GAP) layer. These layers are connected to four Fully-Connected (FC) layers: the first three layers have 2,048, 1,024, 256 units respectively, the output layer has either one output unit to predict MOS, or five units to predict distributions of ratings.

All the three FC layers used the Rectified Linear Unit (ReLU) as the activation function and were followed by a dropout layer, each with rates of 0.25, 0.25, and 0.5 in order to avoid over-fitting. The final layer was linear when we predict MOS directly, and the soft-max activation was used when we predict the distribution of ratings.


\subsection{Loss functions}
\label{subsec:loss}
In machine learning, a loss function defines the "cost" associated with a wrong prediction. Viewing training as an optimization problem, we seek to minimize the associated loss function. We evaluated the performance of five loss functions. For MOS prediction we used Mean Absolute Error (MAE loss) \cite{conv1,boss} and Mean Squared Error (MSE loss) \cite{li2016no}. When we predicted the distribution of ratings, we used cross-entropy loss, Huber loss \cite{hubertus}, and the Earth Mover's Distance (EMD) loss \cite{rubner2000earth}.

\subsubsection{Predicting MOS}
Let $(\textbf{x}, q)$ be the training data, where $\textbf{x}$ is the input image, $q$ is the MOS of the image $\textbf{x}$. 
Given the predicted MOS $\hat{q}$ acquired by feeding $\textbf{x}$ into our proposed system, the most straightforward way is to compute the MAE as loss function,
$
L_{\text{MAE}}(q,\hat{q}) =|q-\hat{q}|.
$
An alternative loss function is the MSE, 
$
L_{\text{MSE}}(q,\hat{q}) = (q-\hat{q})^2,
$
which is differentiable also at the origin, and thus can produce smoother gradients for small errors than the MAE, but penalizes larger deviations from the ground truth more heavily. 
%
%
%
%


\subsubsection{Predicting distribution of ratings}

To predict the distribution of ratings, we denote the input training data by $(\textbf{x}, \textbf{p})$, where $\textbf{x}$ is the input image, and $\textbf{p} = (p_1, \ldots, p_N)$ is its distribution of ratings. We used 5-point ACR for subjective quality assessment of KonIQ-10k, so here $N=5$.
Given the predicted distribution of ratings   $\hat{\textbf{p}} = (\hat{p}_1, \ldots, \hat{p}_N)$, its predicted MOS can be estimated as
\begin{equation}\nonumber
\mathrm{MOS}(\hat{\textbf{p}}) = \sum_{n=1}^N n \cdot \frac{\hat{p}_n}{\hat{p}_1 + \cdots + \hat{p}_N}.
\end{equation}

For image classification, the cross-entropy loss is a standard. We can use the same loss definition for our regression problem:
%
%
%
%
\begin{equation}\nonumber
    L_{\text{cross-entropy}}(\textbf{p},\hat{\textbf{p}})= 
        -\sum_{n=1}^N p_n \log\hat{p}_n.
\end{equation}

\noindent Huber loss for a scalar prediction error $x$ is defined by:
\begin{equation}\nonumber
h_{\delta}(x)=
\begin{cases}
\frac{1}{2}x^2 & |x|\leq \delta, \\
\delta \cdot (|x|-\frac{\delta}{2}) & \text{otherwise,}
\end{cases}
\end{equation}
where $\delta>0$ controls the degree of influence given to larger prediction errors. We chose $\delta=\frac{1}{9}$, the same as in \cite{varga2018deeprn}. 
%
%
%
%
As a result, the Huber loss for predicting distributions of ratings is given as:
$
L_{\delta}(\textbf{p},\hat{\textbf{p}})=\sum_{n=1}^N h_{\delta}(p_n-\hat{p}_n).
$



\begin{table}[tbp]
\setlength\extrarowheight{2pt}
\centering
\begin{tabular}{l r r } \hline
CNN body & Depth & Parameters ($10^6$) \\ \hline
VGG16 \cite{vgg}& 19 & 18.1 \\
ResNet101 \cite{he2016deep}& 489& 49.2\\
InceptionV3 \cite{szegedy2016rethinking}& 311& 28.4\\
InceptionResNetV2 \cite{szegedy2017inception} & 780 & 59.8\\
NASNetMobile \cite{zoph2017learning}& 757& 8.8 \\ \hline
\end{tabular} 
\caption{The CNNs selected for transfer learning, where the depth excludes top layers.}\label{tb:cnn} 
\vspace{-10pt}
\end{table}

Talebi et al.~\cite{talebi_nima_2017} introduced the Earth Mover's Distance (EMD) loss in their work on BIQA, suggesting an improved performance over cross-entropy. The loss is defined as the root mean squared difference between the predicted and ground truth cumulative distributions of scores.
The EMD loss for a predicted distribution \(\hat{\textbf{p}} = (\hat{p}_1, \ldots, \hat{p}_N)\) with cumulative distribution $\textbf{c}_{\hat{p}}$ and ground truth distribution $\textbf{p} = (p_1, \ldots, p_N)$ with cumulative distribution $\textbf{c}_p$ is
\begin{equation}\nonumber
L_{\text{EMD}}(\textbf{p},\hat{\textbf{p}})=
\left(\frac{1}{N}\sum_{n=1}^N (c_{p,n}-c_{\hat{p},n})^2\right)^{1/2}.
\end{equation}

\section{Experimental results}

We evaluated our proposed deep BIQA model on two benchmark databases: the one  proposed in this article, namely KonIQ-10k, and the other was LIVE in the Wild (LIVE-itW) \cite{ghadiyaram:2016massive}.

\subsection{Setup}
To make a fair and comprehensive comparison, we carried out our experiments as follows: We divided KonIQ-10k into three sets, a training set (7,058 images), a validation set (1,000 images), and a test set (2,015 images). The training set was used to train our model, the validation set was used to find the best generalizing model, and the test set was used to evaluate the final model performance that was reported.
Similar to previous works, we used two metrics to evaluate our BIQA methods: the Spearman Rank Order Correlation Coefficient (SROCC) and the Pearson Linear Correlation Coefficient (PLCC). 
We fined-tuned five state-of-the-art CNNs, see Table~\ref{tb:cnn}, each with all five loss functions given in Subsection \ref{subsec:loss}. All the CNN base models were initialized with pre-trained weights from ImageNet. The weights of the top fully-connected layers were initialized using the method of He et al.~\cite{henormal}. 

In all experiments, we used the Adam \cite{kingma2014adam} optimizer, with the default parameters $\beta_1 = 0.9$, $\beta_2 = 0.999$ and a custom learning rate $\alpha$. We first set the learning rate $\alpha = 10^{-4}$ and trained for 40 epochs. In the training process, monitoring the PLCC on the validation set and saving the best performing model. On completion of the initial 40 epochs, we loaded the best model. We ran another 20 epochs with a lower learning rate $\alpha = 5\times10^{-5}$, and subsequently followed the same procedure for another ten epochs with learning rate $\alpha = 10^{-5}$.

All the models and loss functions were implemented using the Python Keras library with Tensorflow as a backend \cite{keras} and ran on two NVIDIA Titan Xp GPUs. The CNN models we used have different numbers of layers and parameters, so it was not possible to train all models with the same batch size in our experiments. We used the largest batch size of $2, 4, 8, 16, \ldots$ images that fit on the available GPU memory for each model.

\subsection{Performance evaluation and discussion}

%
%

\begin{figure*}[!ht]
\centering
\begin{minipage}{0.4\linewidth}
\centerline{\includegraphics[width=\textwidth,height=140pt]{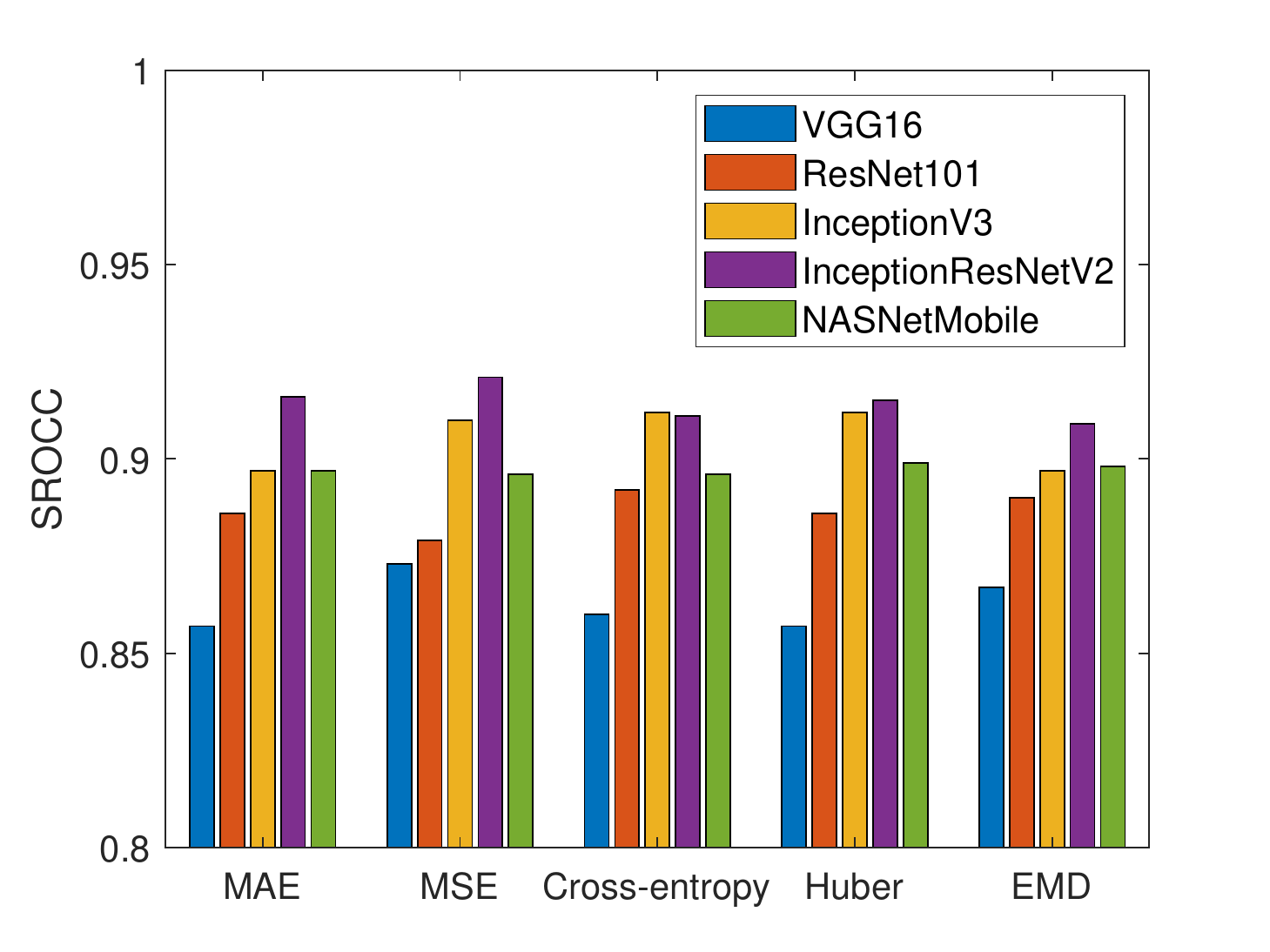}}
\centerline{(a) Trained and tested on KonIQ-10k}
\end{minipage}
\begin{minipage}{0.4\linewidth}
\centerline{\includegraphics[width=\textwidth,height=140pt]{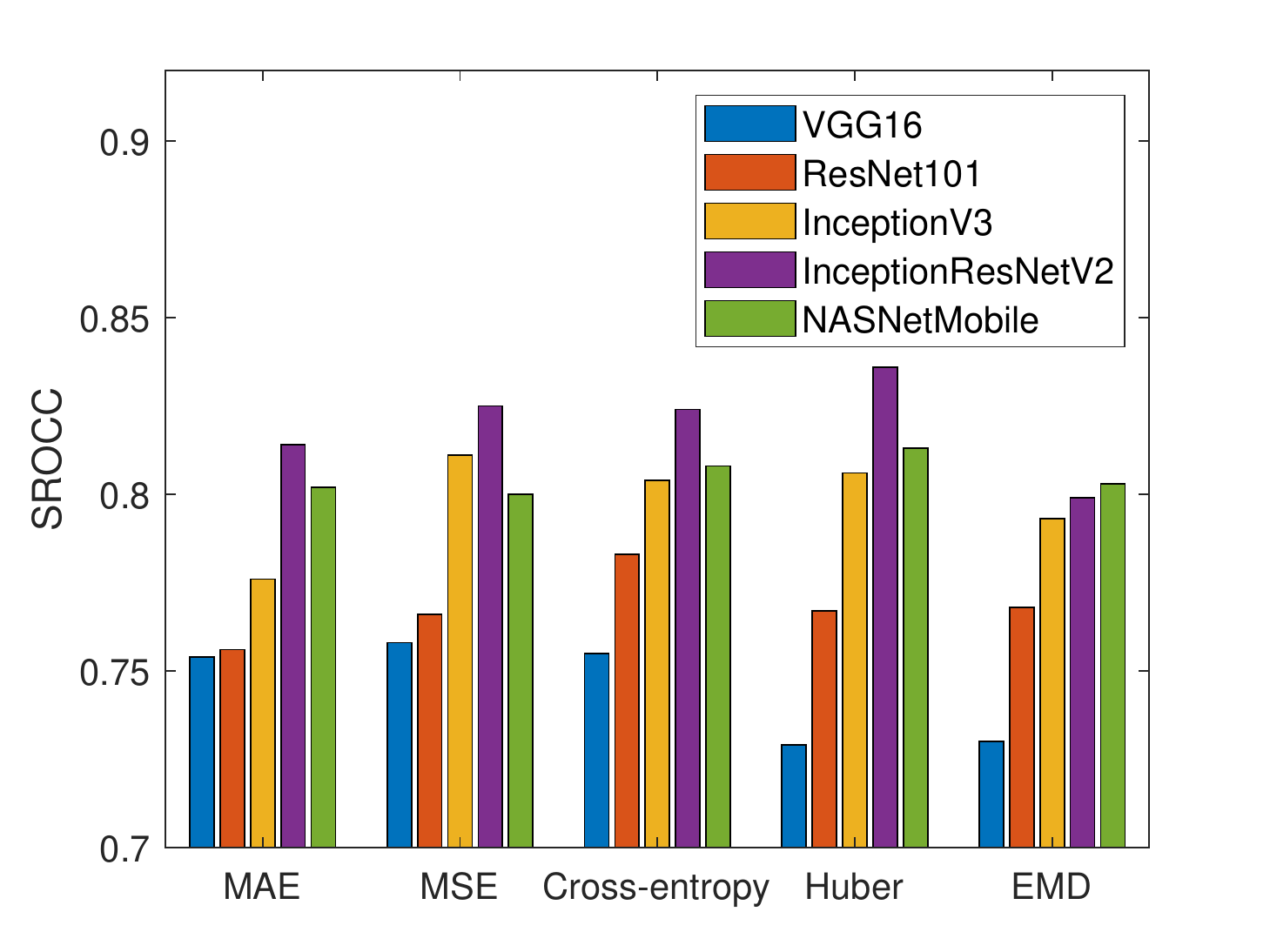}}
\centerline{(b) Trained on KonIQ-10k, tested on LIVE-itW}
\end{minipage}
\caption{Loss function and base architecture comparison at $512\times384$ px, trained on KonIQ-10k training set, tested on (a) KonIQ-10k test set and (b) entire LIVE-itW.}
\label{fig:512-test}
\vspace{-15pt}
\end{figure*}

\begin{figure}[!t]
\centering
\vspace{-5pt}
\includegraphics[width=0.8\linewidth,height=140pt]{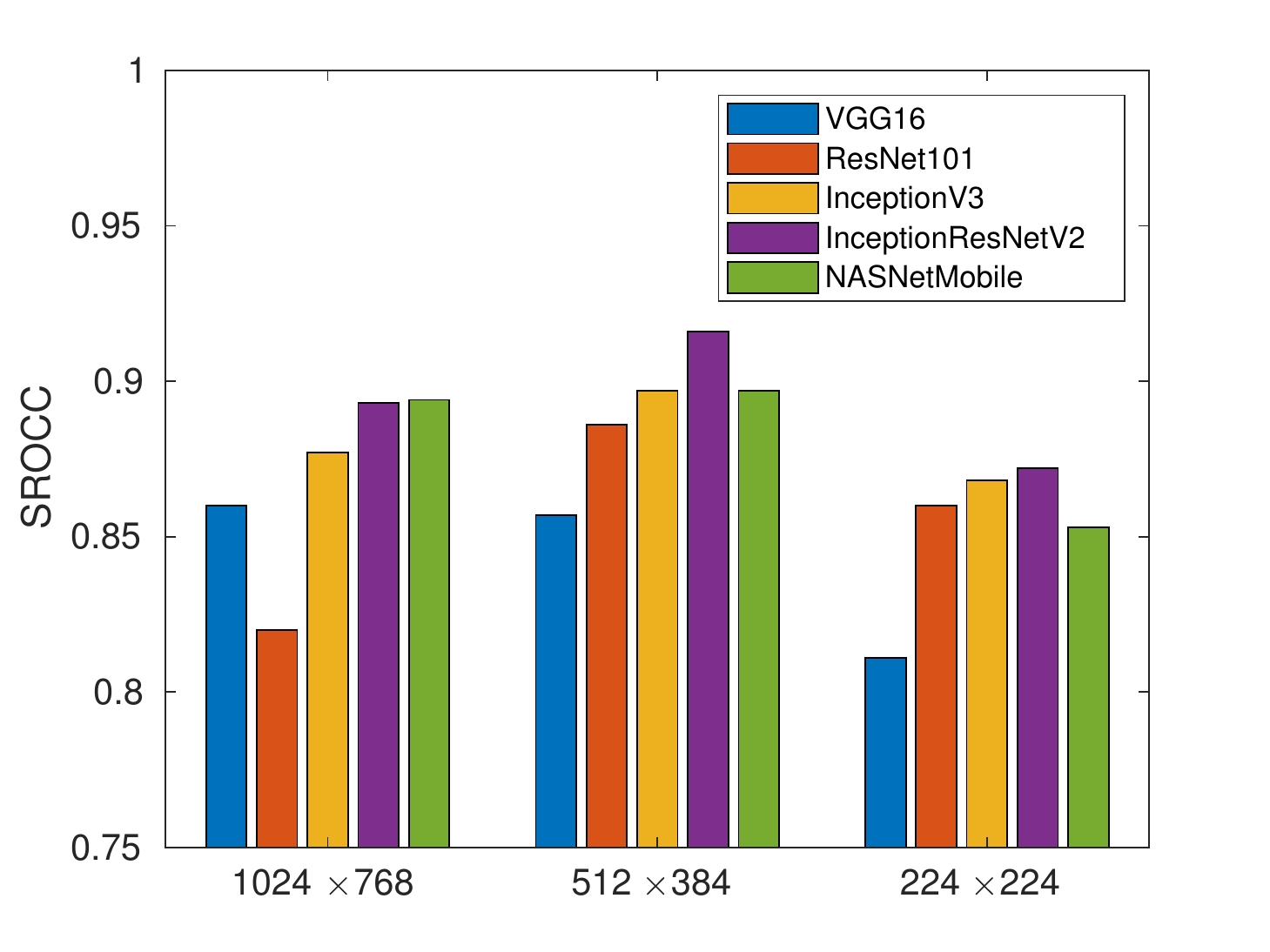}
\caption{Resolution comparison on KonIQ-10k with MAE loss.}
\label{fig:resolution-test-koniq}
\vspace{-0pt}
\end{figure}

\subsubsection{Best model selection: KonCept512}
We considered three factors to establish the best model: input resolution, the loss functions, and the CNN base architectures.

We first found the optimal input resolution for our model. For this purpose, we trained models using the original resolution ($1024 \times 768$) and two down-sampled resolutions ($512 \times 384$ and $224 \times 224$). 
With the default hyper-parameters for training, the results with the MAE loss are shown in Fig.~\ref{fig:resolution-test-koniq}.
Models that were trained on the smallest input size ($224 \times 224$)  had a lower performance than all the others. This suggests that much of the quality-related information was lost during the down-sampling process. Surprisingly, all of the models -- trained on half-sized inputs -- performed better than those trained with original size, except VGG16. Initially, we expected that the performance would improve with more information being available during training. 

A possible reason for the better performance on $512 \times 384$ was that all CNN base architectures had been optimized for small resolution images, such as $224\times 224$ or $512 \times 384$ pixels, and did not perform well for much larger input sizes.  Another reason for this could be that training with very small batch sizes limits the best possible performance that could be achieved \cite{keskar2016large}. For some models, we were only able to use a maximum batch size of 4 when training on $1024 \times 768$ input images. We have not been able to confirm this hypothesis. However, later experiments -- presented in Table~\ref{table:features-retraining} -- suggest this to be the case. There, we were able to train on content features from pre-trained ImageNet architectures extracted from $1024 \times 768$ input images, and these performed better than the features extracted from lower resolutions.

With the optimal input size of $512 \times 384$, we evaluated the performance of the five CNN base architectures with five loss functions. 
Apart from testing them on the KonIQ-10k test set, we also cross-tested them on the entire LIVE-itW database. The SROCC measures are reported in Fig.~\ref{fig:512-test}.
Our analysis showed that all tested CNN base architectures worked well for BIQA. ``Deeper" architectures performed better, with InceptionResNetV2 achieving the best performance. Although the MSE-loss performed the best among the five losses on the KonIQ-10k test set, the improvement was marginal compared to MAE and Huber losses.

Huber loss, applied to distributions, achieved the best performance (0.836 SROCC) when cross-tested on LIVE-itW. 
Furthermore, all models trained on KonIQ-10k had similar low performance when cross-tested on LIVE-itW (about 0.1 SROCC lower on LIVE-itW compared to KonIQ-10k test set), even though their performances on KonIQ-10k were substantially different.

We named our best performing model as KonCept512. It applies the InceptionResNetV2 base architecture, with the MSE loss, and was trained and tested on downscaled $512\times 384$ images. LIVE-itW images were resized to this resolution (from the original $500\times 500$), which ensured the best cross-test performance in contrast to running the model on their original resolution.

\subsubsection{Comparison with state-of-the-art BIQA methods}

We compared KonCept512 with the state-of-the-art BIQA methods, both feature-based and deep-learning-based. We collected seven conventional BIQA methods, with the source code made available by their respective authors. These methods were BIQI \cite{Moorthy:2010}, BLIINDS-II \cite{blind2}, BRISQUE \cite{bris}, CORNIA \cite{ye2012unsupervised}, DIIVINE \cite{Moorthy:2011}, HOSA \cite{hosa}, and SSEQ \cite{Liu:2014b}.
For conventional BIQA methods, we used an SVR (RBF kernel) to train and predict the quality from extracted handcrafted features. 
For deep-learning-based BIQA methods, we reimplemented BosICIP \cite{boss}, CNN \cite{conv1}, and DeepBIQ \cite{bianco2018use}. BosICIP and CNN were trained from scratch, and DeepBIQ was trained on the pre-trained deep network backbone on ImageNet. 
All of them were trained only on the training set of KonIQ-10k and tested both on the test set of KonIQ-10k and the entire LIVE-itW database.

TABLE~\ref{tab:perfcomp_koniq} presents the results. The performance of conventional BIQA methods on KonIQ-10k was far from satisfactory, even if they had achieved promising performances on artificially distorted databases. The methods based on local features, such as CORNIA and HOSA, showed a better performance than those based on global features.

Further, regarding the deep-learning-based methods, since both BosICIP and CNN are very ``shallow" convolutional neural network architectures (trained from scratch), their performance was lower than most conventional BIQA methods and lower than our proposed model.
%
%
%
%
Bianco et al.\ \cite{bianco2018use}, in their work on DeepBIQ, relied on a base CNN architecture inspired by AlexNet. It predicts the overall quality of an image as the average of local patch-wise quality. In our re-implementation, we used the better-performing InceptionResNetV2 architecture, as well as VGG16, with $224\times224$ crops taken from the half-sized images of KonIQ-10k ($512\times384$). This resolution gave a better performance than extracting patches from the original images (SROCC/PLCC on the KonIQ-10k test set was 0.890/0.892 and on LIVE-itW: 0.769/0.808). The predictor we used was the same three-layer fully-connected head as in the rest of our experiments with an MSE loss, predicting MOS values.

\begin{table}[t]
\setlength\extrarowheight{2pt}
\caption{Comparison to state-of-the-art methods, trained on KonIQ-10k and tested on both KonIQ-10k and LIVE-itW.}\label{tab:perfcomp_koniq} \centering
\resizebox{0.5\textwidth}{!}{
\begin{threeparttable}
\begin{tabular}{l|c c|c c}
  \hline
  &\multicolumn{2}{c|}{\textbf{Test on KonIQ-10k}}&\multicolumn{2}{c}{\textbf{Test on LIVE-itW}} \\
  \textbf{Method} & \textbf{SROCC} & \textbf{PLCC}& \textbf{SROCC} & \textbf{PLCC}\\ \hline \hline
  BIQI &  0.559&0.616 &0.364&0.447\\
  BLIINDS-II &0.585&0.598&0.090 &0.107 \\
  BRISQUE & 0.705&0.707&0.561 & 0.598\\
  CORNIA\tnote{*} &0.780&0.808&0.621&0.644\\
  DIIVINE & 0.589&0.612&0.435&0.478 \\
  HOSA\tnote{*} &0.805&0.828 &0.628 &0.668 \\
  SSEQ &0.604&0.612&0.245&0.286 \\
  \hline
  BosICIP      &  0.604    & 0.606 &0.493&0.482\\
  CNN          &  0.572    & 0.584& 0.465&0.450\\
  DeepRN (ResNet101) & 0.867 & 0.880 &0.726 &0.750\\ 
  DeepBIQ (VGG16) & 0.872 & 0.886 &0.742 &0.747\\ 
  DeepBIQ (InceptionResNetV2) & 0.907     & 0.911 &0.804&0.821\\ 
  DistNet-Q3 \cite{dendi2018generating} & 0.700 & 0.710 & -- & -- \\ 
  Learning-to-Rank IQA \cite{zhang2019learning} &0.892 & -- & -- & -- \\
  \hline
  KonCept512 &\bf 0.921 &\bf 0.937 &\bf 0.825&\bf 0.848\\
  \hline
\end{tabular}
\begin{tablenotes}
 \item[*] To reduce computational cost, we reduced CORNIA and HOSA feature vector from 20,000 dimensions and 14,700 dimensions respectively to 100 dimensions using PCA. 
\end{tablenotes}
\end{threeparttable}
}
\vspace{-12pt}
\end{table}

\begin{table*}[!ht]
\centering
\setlength\extrarowheight{2pt}
\begin{tabular}{lllllllll} \hline
 & Training set size & 1000  & 2000  & 3000  & 4000  & 5000  & 6000  & 7000  \\ \hline
\multirow{2}{*}{SROCC on KonIQ-10k (test)} & Mean        & 0.831 & 0.871 & 0.889 & 0.899 & 0.908 & 0.912 & 0.915 \\
                                & SD          & 0.013 & 0.010 & 0.013 & 0.007 & 0.005 & 0.007 & 0.006 \\ \hline
\multirow{2}{*}{SROCC on LIVE-itW (all)}       & Mean        & 0.757 & 0.791 & 0.804 & 0.813 & 0.818 & 0.822 & 0.827 \\
                                & SD          & 0.009 & 0.011 & 0.011 & 0.005 & 0.005 & 0.004 & 0.005 \\ \hline
\end{tabular}
\caption{Effect of changing the training/validation/test split. The results are based on repeating the training on 10 additional random splits of the KonIQ-10k database (1000 images for validation and 2000 for testing). The mean and standard-deviation (SD) of the SROCC to the ground-truth MOS is shown for each training set size.}
\label{tab:multiple-training-splits}
\vspace{-10pt}
\end{table*}

With the help of ``deeper'' architectures, the performance of DeepBIQ (InceptionResNetV2) increased by more than 0.1 compared to the best conventional BIQA methods. By training and testing on entire images to preserve content information, KonCept512 improved the SROCC by around 0.02 on both sets compared to the local patch-based DeepBIQ (InceptionResNetV2). Scatterplots of the IQA predictions from KonCept512 and the ground truth MOS are presented in Fig.~\ref{fig:512-test-plot}. Some prediction examples from the KonIQ-10k test set are shown in Fig.~19 (Supplementary Material).

In another previous work, Varga et al.\ \cite{varga2018deeprn} proposed the DeepRN architecture based on ResNet101, which showed promising results on KonIQ-10k. 
For a better comparison of our work, we reimplemented DeepRN in our framework. The performance on the train/validation/test split used throughout our experiments was lower than previously reported. In the original experiments in \cite{varga2018deeprn}, the authors used a different parameterization of the training procedure. While we consistently used a fixed validation set, DeepRN in \cite{varga2018deeprn} did not always follow this procedure. Moreover, the choice of the train/test split was different.

For the cross-test on the LIVE-itW database, the performance was lower than for the test set in KonIQ-10k (SROCC of 0.825 versus 0.921, see Table~\ref{tab:perfcomp_koniq}). This may be due to the different ways the images were collected in the two databases. In KonIQ-10k, the images were cropped from the originals as captured by cameras in order to maintain their original aspect ratio (and quality), whereas, in LIVE-itW, images were rescaled changing their original aspect-ratio, stretching them unevenly. This type of defect was improbable to happen in images that were selected as part of KonIQ-10k, and thus introduced a type of degradation that our trained models were not aware of.

It should be noted that deep-learning-based methods generalized better than conventional methods when cross-tested, i.e., trained on KonIQ-10k and tested on LIVE-itW. When cross-testing conventional methods, they exhibited substantial performance drops, whereas the correlation decrease for deep-learning-based methods was only around 0.1 SROCC.

\subsubsection{Effects of the training set size}
The size of the training set can be expected to have a substantial effect on the performance of the machine learning system. This had been the case for image classification, where the training datasets contained millions of images \cite{deng2009imagenet}. For BIQA, we only trained on 7,000 images, so we studied the relationship between training set size and test performance in order to extrapolate it to larger training set sizes.

\begin{figure}[!t]
\vspace{-10pt}
\centering
\includegraphics[width=0.8\linewidth,height=140pt]{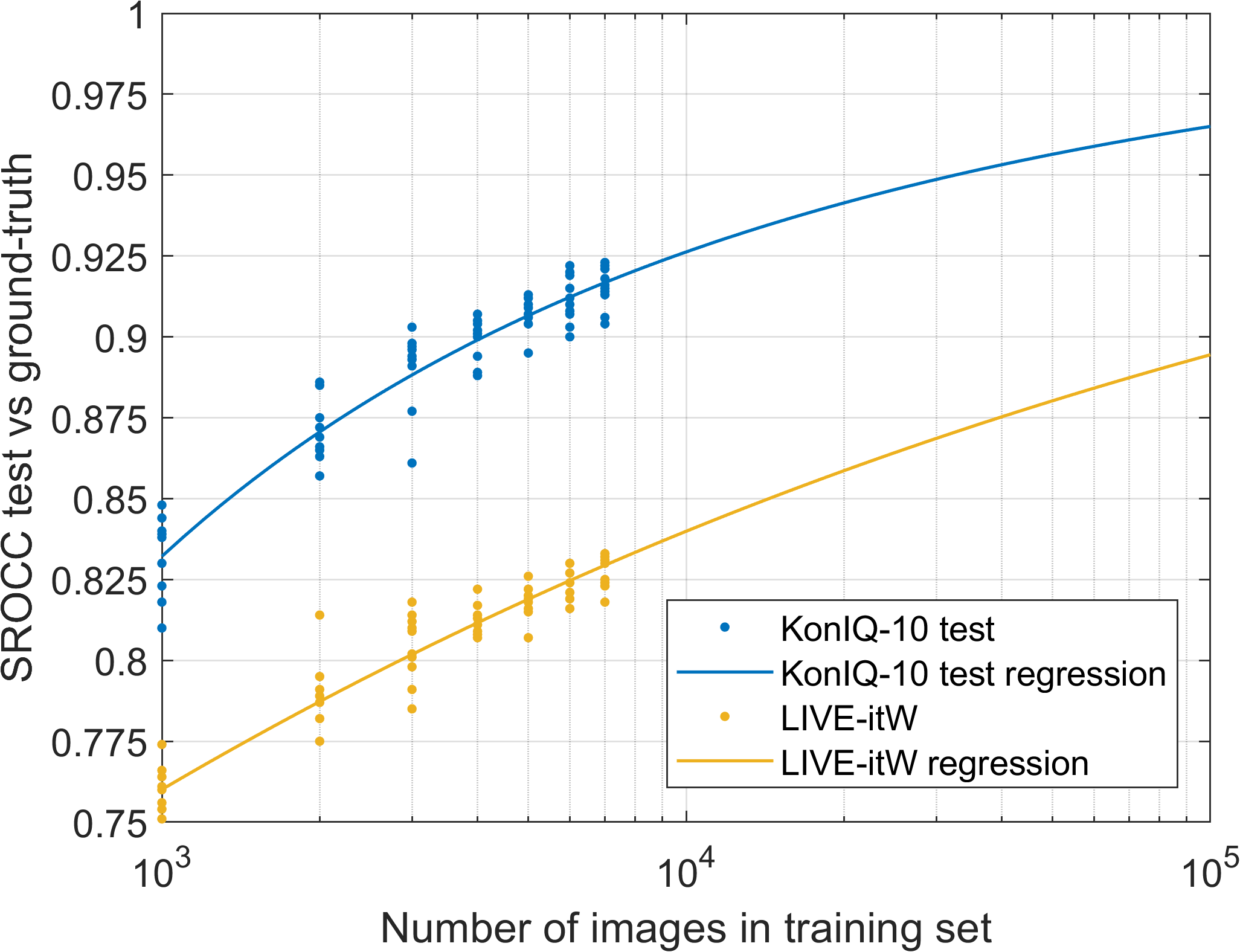}
\caption{Performance of KonCept512 on the KonIQ-10k test set and the entire LIVE-itW database, when the training set size from KonIQ-10k is increased from 1,000 to 7,000 items. Multiple samples of training, validation, and test sets are used. For each re-sample, when training on a larger set size, items previously used from the smaller sets are included. A curve is fitted to the scatter plots, to extrapolate the performance of the model to larger training sizes. For more details, see Table \ref{tab:multiple-training-splits}.}
\label{fig:extrapolation-live}
\vspace{-15pt}
%
%
%
%
\end{figure}
%
%
%
%
We ran the training procedure using KonCept512 with several intermediate training sizes ($x = 1,000, 2,000, \ldots, 7,000$ images), validated, and tested the performance using the original validation and test set from KonIQ-10k. An additional ten resamples were created, consisting of random splits for training (of size $x$), validation (1000 images), and test sets (2000 images). Moreover, we cross-tested each retrained model on the LIVE-itW database.

To extrapolate the SROCC performance for larger training-set sizes, we fitted the function $f(x) = 1 - (x^a + b)^{-1}$ to the multiple data points at training sizes $x$ from 1,000 to 7,000 images. This function was chosen for its simplicity (having only two parameters) to avoid over-fitting. Moreover, it has the desired properties of monotonicity and convergence to an SROCC of 1.0 as the training set size $x \rightarrow \infty$. The results are presented in Fig.~\ref{fig:extrapolation-live}.  The function type estimates the test performance on KonIQ-10k well.

The extrapolated performances when $x = 100,000$ ``virtual'' training images would be used, similar to those in KonIQ-10k, are $0.965\pm0.025$ SROCC w.r.t.\ the ground truth KonIQ-10k test data, and $0.895\pm0.021$ SROCC w.r.t.\ cross-test on LIVE-itW. We computed 95\% observational confidence bounds by bootstrapping, giving  $\pm0.025$ and $\pm0.021$, respectively. 

Crowdsourcing experiments in \cite{QoMEXReliability,ponomarenko:2009tid2008} exhibited SROCC agreements with a range of the following $[0.91,0.96]$ between multiple repeats of the same experiment, depending on the source and number of participants. For KonIQ-10k, the mean agreement goes up to $0.973$ (SROCC) when half of the participants' is compared to the other half's, as mentioned in Section \ref{sec:reliability}, see Fig.~\ref{fig:group-agreement}. 



\subsubsection{On the prediction power of IQA methods}

The performance of IQA methods usually is assessed by SROCC values w.r.t.\ some benchmark IQA dataset. In our case, the quality predictions of KonCept512 on the test set of KonIQ-10k yielded an SROCC of 0.921 with the MOS values from 120 votes per image, see Table \ref{tab:perfcomp_koniq}. While SROCC values are useful when comparing competing IQA methods, they are difficult to interpret intuitively as absolute measures. How good is an SROCC on a particular test dataset, for example, 0.921 on KonIQ-10k? In this subsection, we provide such an intuitive understanding. 

Assuming a given ground truth for the image quality values in a test dataset, we may compare the performance in terms of SROCC, of an objective IQA method with that of a random group of $N$ users, each one providing one judgment for every stimulus in the test set. The MOS values of the group will give rise to some SROCC w.r.t.\ the ground truth values. Moreover, it is expected that, on average, the SROCC is monotonically increasing with the group size $N$ (if sufficiently many SROCC results are averaged). We may now ask for the maximal group size $N_\text{max}$, for which the SROCC does not exceed that which is provided by the objective IQA method. The interpretation than would be that the proposed objective IQA method gives results comparable to a group of $N_\text{max}$ judges. In the following, we will argue that with KonCept512 applied to the test set of KonIQ-10k, we have $N_\text{max} \approx 9$, so KonCept512 provides the results that correlate with the ground truth as well as nine randomly chosen votes per stimulus.

To carry out the program outlined in the previous paragraph, we need to define the ``ground truth'', and then randomly sample a group of users providing $N$ ratings for each test set item. We have a total of 1,459 workers that have provided scores, however, for differing subsets of images. To define the ground truth, we have to rely on these scores, and we must take care that when we sample groups of users that these do not overlap with the workers who provided the scores for the ground truth. Therefore, we randomly sampled half of the participants and regarded their corresponding MOS values as ground truth. This resulted in $\approx57$ votes per image on average. Thus, $N=57$ can be regarded as the ``effective group size''. We have done 200 random bootstraps and found a very high average SROCC of $0.973$ between their MOS values, see Fig.\ \ref{fig:group-agreement}. Therefore, we think that 57 crowdsourced votes suffice to define the ground truth.

\begin{figure*}[!ht]
\centering
\begin{minipage}{0.4\linewidth}
\centerline{\includegraphics[width=\textwidth]{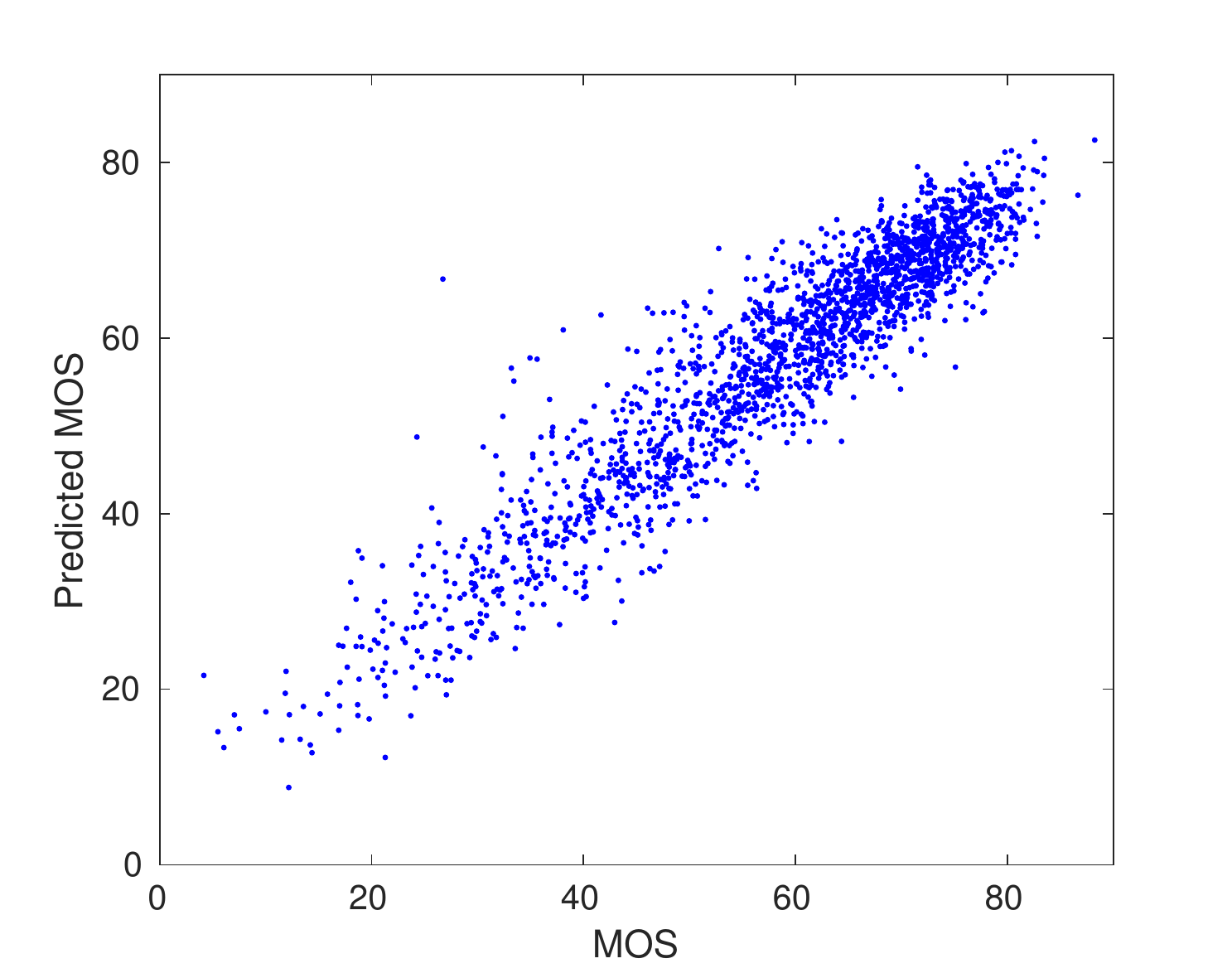}}
\centerline{(a) Trained and tested on KonIQ-10k}
\end{minipage}
\begin{minipage}{0.4\linewidth}
\centerline{\includegraphics[width=\textwidth]{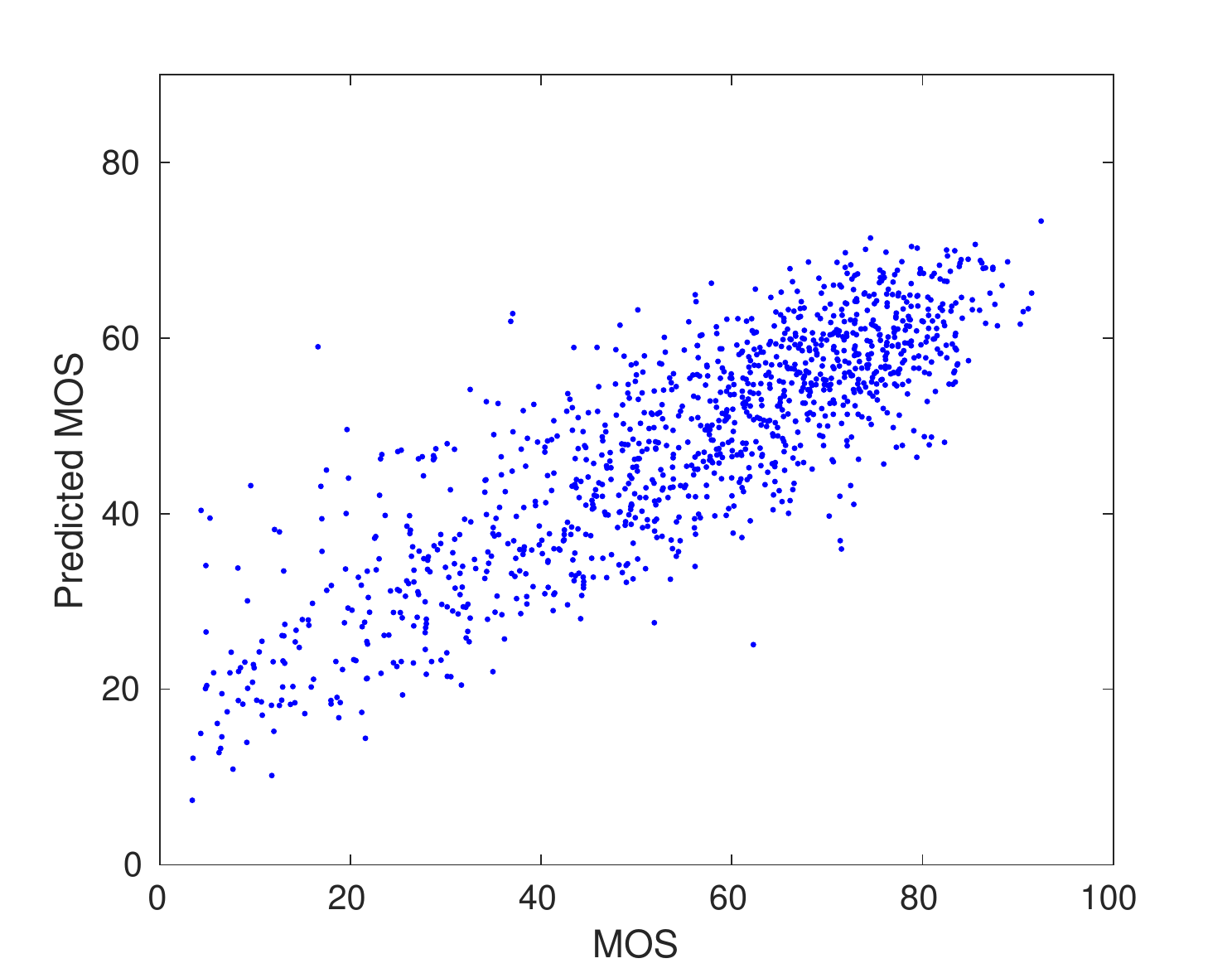}}
\centerline{(b) Trained on KonIQ-10k, tested on LIVE-itW}
\end{minipage}
\caption{Scatter plots of predicted MOS by KonCept512 versus ground truth MOS. The model was trained on KonIQ-10k training set (7,058 images), tested on (a) KonIQ-10k test set (2,015 images) and (b) Entire LIVE-itW (1,169 images). In the scatter plots, each point corresponds to an image, the x-axis denotes the ground truth MOS obtained from crowd workers, and y-axis denotes the predicted MOS.
}
\label{fig:512-test-plot}
\end{figure*}



\begin{figure}[!t]
\centering
\vspace{-0pt}
\includegraphics[width=0.8\linewidth]{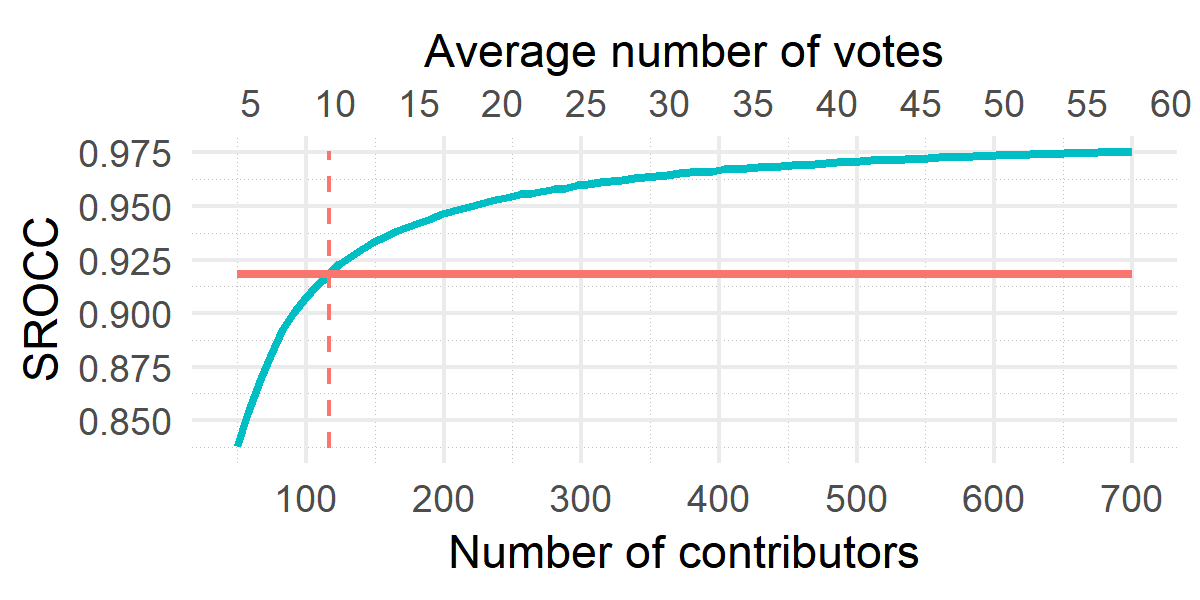}
\caption{Cyan curve: mean SROCC between MOS values from random groups (A) of increasing numbers of contributors and non-overlapping random groups of 700 contributors (B). SROCC is computed on the KonIQ-10k default test set of 2015 images. Red line: 0.918 mean SROCC between our KonCept512 predictions on the KonIQ-10k test set, and the MOS from 700 randomly chosen contributors. KonCept512 performs similar to MOS values obtained from 9 contributor votes per image.}
\label{fig:group-agreement-2}
\end{figure}

\begin{figure*}[!t]
\centering
\includegraphics[width=0.7\linewidth]{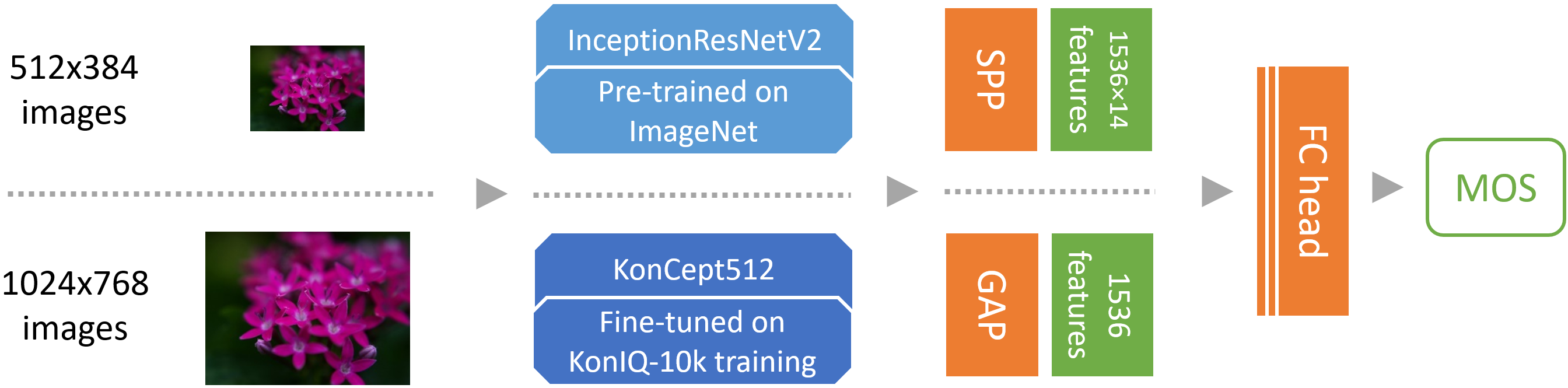}
\caption{Choices made when training on features: 1. extracted from either half or full-sized images from KonIQ-10k, 2. features coming from either pre-trained or fine-tuned InceptionResNetV2 base network, with 3. SPP[1,2,3] or GAP pooling.}
\label{fig:feature-learning}
\end{figure*}

\begin{table*}[htb]
\centering
\setlength\extrarowheight{3pt}
\resizebox{\textwidth}{!}{%
\begin{tabular}{l|ccccccccccc}
\cline{2-6} \cline{8-12}
\multicolumn{1}{c|}{} & \multicolumn{5}{c|}{Deep IQA features, from KonCept512 trained on KonIQ-10k at $512\times384$} & \multicolumn{1}{c|}{} & \multicolumn{5}{c|}{Content features, from InceptionResNetV2 pretrained on ImageNet} \\ \cline{2-6} \cline{8-12} 
\multicolumn{1}{c|}{} & \multicolumn{2}{c|}{$512\times384$} & \multicolumn{1}{c|}{} & \multicolumn{2}{c|}{$1024\times768$} & \multicolumn{1}{c|}{} & \multicolumn{2}{c|}{$512\times384$} & \multicolumn{1}{c|}{} & \multicolumn{2}{c|}{$1024\times768$} \\ \cline{2-3} \cline{5-6} \cline{8-9} \cline{11-12} 
 & \multicolumn{1}{c|}{KonIQ-10k} & \multicolumn{1}{c|}{LIVE-itW} & \multicolumn{1}{c|}{} & \multicolumn{1}{c|}{KonIQ-10k} & \multicolumn{1}{c|}{LIVE-itW} & \multicolumn{1}{c|}{} & \multicolumn{1}{c|}{KonIQ-10k} & \multicolumn{1}{c|}{LIVE-itW} & \multicolumn{1}{c|}{} & \multicolumn{1}{c|}{KonIQ-10k} & \multicolumn{1}{c|}{LIVE-itW} \\ \cline{1-3} \cline{5-6} \cline{8-9} \cline{11-12}  
\multicolumn{1}{|c|}{SPP} & \multicolumn{1}{c}{0.912 / 0.930} & 0.831 / 0.846 &  & 0.913 / 0.928 & 0.824 / 0.839 & \multicolumn{1}{c}{} & \multicolumn{1}{c}{0.650 / 0.709} & 0.592 / 0.622 &  & 0.781 / 0.819 & 0.597 / 0.633 \\ \cline{1-1}
\multicolumn{1}{|c|}{GAP} & \multicolumn{1}{c}{0.916 / 0.933} & 0.831 / 0.848 &  & 0.918 / 0.930 & 0.815 / 0.830 & \multicolumn{1}{c}{} & \multicolumn{1}{c}{0.668 / 0.719} & 0.582 / 0.617 &  & 0.788 / 0.824 & 0.610 / 0.644 \\ \cline{1-1}
\end{tabular}%
}
\caption{Re-training on KonIQ-10k features: in the table on the left the features are extracted using our best performing KonCept512, and on the right content features are extracted using a pre-trained InceptionResNetV2 on ImageNet. The BIQA network KonCept512 is trained at $512\times384$. Features are extracted from images at both resolutions $512\times384$ and $1024\times768$. Two types of features are considered: Spatial Pyramid Pooling (SPP), with average pooling and sizes ${1,2,3}$, and Global Average Pooling (GAP). The cross-test on the LIVE-itW database is checks the generalization performance, which is obtained when imagesare resized to $512\times384$. The training loss used in all cases is the Mean Squared Error and we predict MOS. SROCC/PLCC correlations are shown in each case: for the test set from KonIQ-10k and the entire LIVE-itW database. In the bottom left corner of the table, the SRCC of KonCept512 is slightly lower (0.916) than the best reported (0.921) even if the architectures are identical. This happens due to a different random initialization of the FC head during feature learning.}
\label{table:features-retraining}
\end{table*}

The other half of the participant group served as an independent source for a group sample to be compared with the MOS values of the first half, yielding a corresponding SROCC value. For example, sampling 300 contributors gave rise to an effective group size, $N \approx 25$ respectively, to an average of $N \approx 25$ judgments per image. Finally, the whole sampling procedure for half of the participants for the ground truth and the group from the other half and the SROCC computation was repeated 200 times, and the SROCC values were averaged.

The results for increasing group sizes are given in Fig.\ \ref{fig:group-agreement-2} (cyan curve). Moreover, the average SROCC between the scores of KonCept512 and the ground truth MOS from a random half of all participants is 0.918  (red line in Fig. \ref{fig:group-agreement-2}). The intersection of the two lines corresponds to the average number of scores per image required to achieve the same performance as KonCept512. Thus $N_\text{max} \approx 9$. Hence, KonCept512, applied to the KonIQ-10k test set, is as powerful as groups of $9$ workers, providing a total of 9 ratings for each test image!

\subsubsection{Training on features}

As we did not have the resources to run large batch sizes on images at the original resolution available in KonIQ-10k ($1024\times768$ pixels), we studied the performance of features derived from the best architecture in our experiments: InceptionResNetV2. The base network was trained for BIQA on images with a resolution of $512\times384$ pixels leading to KonCept512. A brief overview of the choices made is presented in Fig.~\ref{fig:feature-learning}.

We extracted two types of IQA features. The first was from the GAP layer of the network (1,536 features), and for the second type, we replaced the GAP layer with a Spatial Pyramid Pooling layer with pooling sizes $\{1,2,3\}$, giving rise to $(1+4+9)\cdot 1,536 =  21,504$ features. For SPP, a pooling size of 1 was the same as the GAP features. We extracted these features both from $512\times384$ and $1024\times768$ input images. With the extracted features, we retrained the same type of head network used in the primary architecture, made of fully connected layers of 2,048, 1,024, and 256 neurons with dropout rates of 0.25, 0.25, and 0.5, respectively. We used the MSE loss to predict mean opinion scores.
%
%
%
%
%

In addition to IQA features, we also extracted content features from the InceptionResNetV2 network, pre-trained on ImageNet, and trained the previously mentioned head-network. The results\footnote{In Table \ref{table:features-retraining} we show the performance when training on features from KonCept512, the same model that was trained on the default train/validation/test split and is compared to in Table \ref{tab:perfcomp_koniq}. In Table \ref{tab:multiple-training-splits} we present the average performance over 11 train/validation/test splits (the default + 10 more) when retraining the KonCept512 model.} are presented in TABLE~\ref{table:features-retraining}, comparing the performance on the KonIQ-10k test set and the performance resulting from cross-testing the trained network on the entire LIVE-itW database.

The best performance on the KonIQ-10k test set was achieved when retraining on GAP features from $1024\times768$ input images. However, the improvement over SPP features was marginal. Moreover, SPP features outperformed GAP features in terms of cross-test performance on LIVE-itW. This suggests that having more information available in the features could help improve generalization. It could prove worthwhile to fine-tune the base architecture on $1024\times768$ images with sufficiently large batch sizes. This is supported by the effect of content features; those extracted from $1024\times768$ outperformed features from $512\times384$ images. 

When training on features approach, there was no improvement over the baseline performance of KonCept512 on KonIQ-10k. There was a slight improvement when generalizing. The cross-test on LIVE-itW improved from an SROCC of 0.825 for KonCept512, to 0.831 for both GAP and SPP features extracted using the KonCept512 network from  $512\times384$ images.

\subsubsection{Ecological validity}

While there is no universally agreed-upon definition of ecological validity, according to Britannica \cite{gouvier_ecological}, it is a measure of how well test performance predicts behaviors in real-world settings, or, in our case, generalization performance of our model to in-the-wild images.

Users from a photography community have common interests, that can be very different w.r.t.\ category, style, and other factors from users from another community, e.g., Flickr.com (mostly amateur photographers) vs.\ 1x.com (professionals). Thus, randomly sampled images from one community are not necessarily representative of another. Ecological validity is tied to a particular community (environment). If we want to devise a database that is more broadly applicable (general ecological validity), we stipulate that a normalization of the distributions of various categories and style attributes has to be done. Our diversity sampling procedure is such a normalization.

From another point of view, machine learning models exhibit an improved generalization performance when trained on representative and balanced datasets. While real-world images are more representative than artificially degraded ones, a balance was ensured via de-duplication and diversity sampling w.r.t.\ the indicators and category features. 

On the one hand, if sample diversity would not have been considered, we would have severely under-sampling extreme attribute values. On the other hand, we might have included too many extreme images relative to the natural distribution of online images. We strike a balance by excluding images at the far ends of each attribute dimension. The cut-offs were decided based on visual inspection.

If one wants to benchmark methods for real-world attribute distributions, then one can repeatedly sample from the dataset using the natural distributions and report the average performance. The natural distribution of the indicators is available with our database for benchmarking purposes.

\section{Conclusion}
\label{sec:conc}

We proposed a new systematic and scalable approach to create KonIQ-10k, the largest ecologically valid IQA database to date. It consists of 10,073 images, more than eight times as many as the state-of-the-art Live-itW. Relying on this dataset, we propose a new deep learning model {KonCept512}, which performs best when tested on KonIQ-10k when compared to existing works, and generalizes very well to LIVE-itW. We have argued that the ability to generalize from KonIQ-10k to LIVE-itW is a consequence of (1) the diversity and representativeness of the training database (KonIQ-10k) for public Internet images, (2) having ran reliable crowdsourcing experiments for both databases, and (3) the technical improvements of our BIQA deep learning architecture, culminating in our best model {KonCept512}.

A more diverse subset of items is more representative of the wider range of images in the wild. Thus, we have ensured KonIQ-10k contains images that are very diverse concerning category, quality-related indicators, and technical parameters, such as compression settings, bitrate, and capture device. For instance, the 10,073 images were taken by 1265 different camera models from about 100 manufacturers.
The experimental design and the post-analysis ensure the quality of the subjective scoring procedure. We validated the quality of the experiments in connection to ratings coming from a panel of photography experts, and have shown a high level of inter-user agreement. Moreover, groups of crowd workers reached a very high inter-agreement -- over 0.97 SROCC -- when at least 57 scores are assigned to each image. Our database provides 120 scores per image. Thus, the MOS scores are precise, having a high agreement when simulating repetitions of the experiment, and are accurate with respect to the opinions of domain experts.

Our best deep model, {KonCept512}, brings several improvements. It relies on the modern InceptionResNetV2 base architecture, which generally performs better than the alternatives tested. There is no clear winner between using the distribution of scores or MOS and among the various losses tested. However, on average, the simple standard losses, like MSE, outperform the ones presented in state-of-the-art deep learning methods, concerning the correlation (SROCC/PLCC) between the predictions and the ground truth MOS. {KonCept512} derives its performance from several other design choices, such as (1) training on larger resolutions ($512\times 384$) than existing works have employed (typically $224\times 224$), (2) choosing the best model that maximizes the correlation to the ground truth, and (3) using a fully connected head architecture for IQA that enables multi-resolution training/testing via a GAP layer.

Overall, the main challenge for the further performance improvement of deep BIQA methods is the size of the training database, given a careful selection of images and their reliable annotation. We predict that datasets with about 100,000 images (built similarly to KonIQ-10k) will close the gap between objective BIQA and the aggregated opinion of large groups of observers in the wild.

\ifCLASSOPTIONcompsoc
  \section*{Acknowledgements}
\else
  \section*{Acknowledgement}
\fi
Funded by the Deutsche Forschungsgemeinschaft (DFG, German Research Foundation) -- Project-ID 251654672 -- TRR 161 (Project A05).
The research was further supported by the Hungarian Scientific Research Fund (No.\ OTKA 120499). We would like to thank Domonkos Varga for his advice about deep learning, in particular for the DeepRN architecture \cite{varga2018deeprn}.
 
\ifCLASSOPTIONcaptionsoff
  \newpage
\fi

\bibliographystyle{IEEEtran}
\bibliography{refs}

\section{Appendix}
\subsection{Tag-based content sampling}

Our heuristic algorithm for the initial tag-based content sampling in Section 3.2 is as follows. Considering the scale of the problem, we propose a computationally efficient method to find an approximate solution. Let $\Phi(t, S_O)$ be the number of images that contain tag $t$ in the set $S_O$ of 4.8 million. We choose a tag quota $Q$ such that all images that contain a tag $t$ with $\Phi(t, S_O) < Q$ are added to the sampled set $S_S$. Let $T(S)$ be the set of tags in a set of images, $S$.  For remaining tags $T_R = T(S_O) \backslash T(S_S)$, we include images in $S_S$ such that at least each tag's quota $Q$ is reached. This procedure is as follows. For each tag $t \in T_R$, in order of increasing counts $\Phi(t, S_O)$, we generate an ordered list of candidate images, $O(S_O \backslash S_S, K_t)$, where the list of images is sorted in decreasing order of  $K_t$, the machine confidence in the presence of the tag $t \in T_R$. $K_t$ is part of the YFCC100m meta-data and is derived from the object classification deep neural network that was used to estimate the most likely tags for each image. Then we add the top $Q - \Phi(t, S_S)$ images from $O(S_O \backslash S_S, K_t)$ to $S_S$. To assure that $|S_S| \approx 1,000,000$, one can apply the bisection method to choose the tag quota $Q$. We ran the above algorithm with $Q = 4000$ and stopped adding images to $S_S$ when $|S_S| = 1,000,000$.

\clearpage
\begin{IEEEbiography}[{\includegraphics[width=1in,height=1.25in,clip,keepaspectratio]{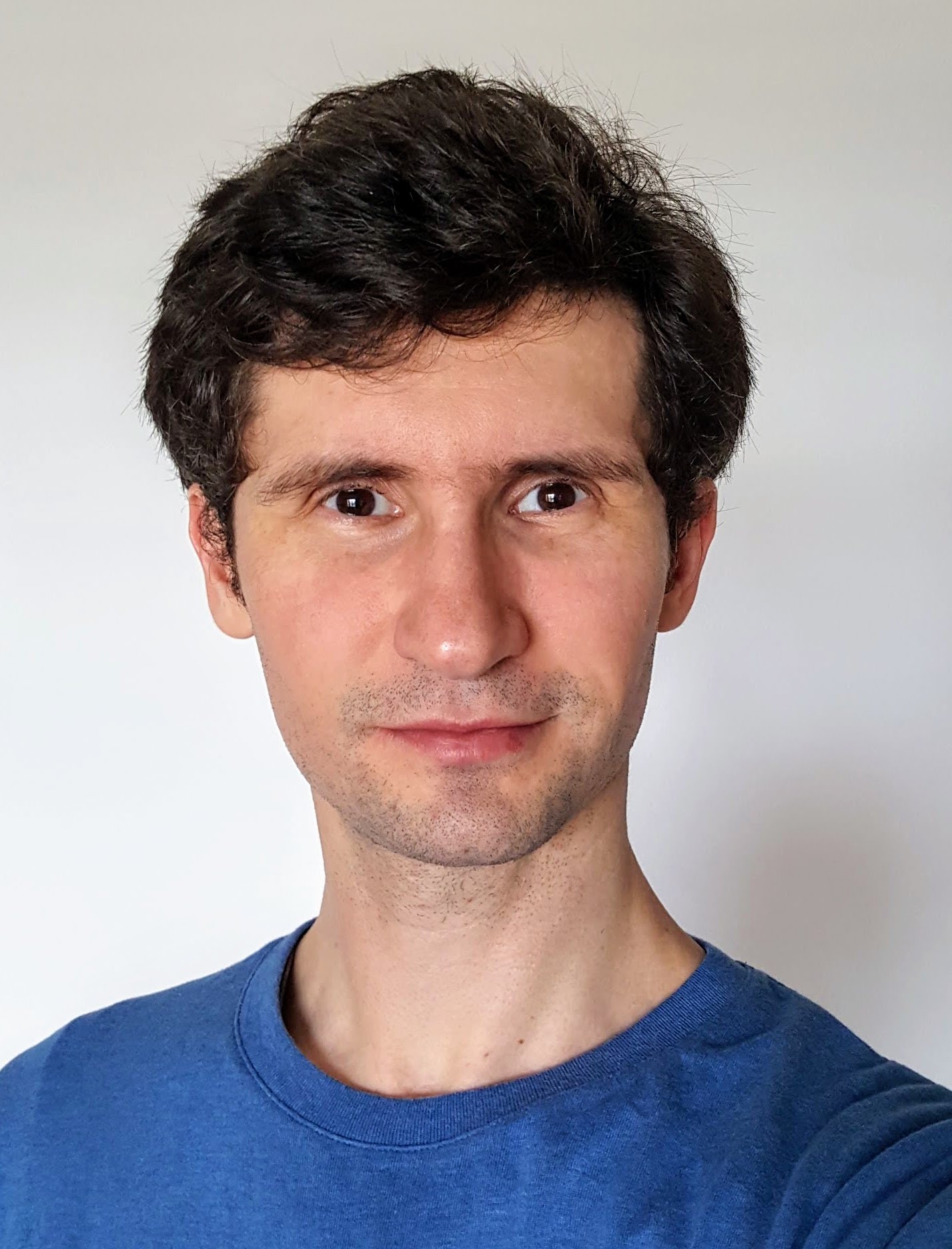}}]{Vlad Hosu} is a postdoc since 2016 at the Department of Computer and Information Science at the University of Konstanz, Germany. Previously he was a Research Fellow at NUS, Singapore, having received his Ph.D. at the same institution in 2014. His research interests include visual quality assessment, image enhancement, effective crowd-sourcing strategies, understanding, and modeling human visual perception.
\end{IEEEbiography}
\vskip 0pt plus -1fil
\begin{IEEEbiography}[{\includegraphics[width=1in,height=1.25in,clip,keepaspectratio]{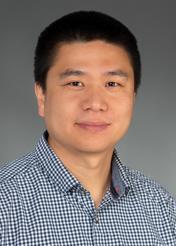}}]{Hanhe Lin} completed his PhD at Department of Information Science, University of Otago, New Zealand in 2016. Now he is a postdoc researcher at the Department of Computer and Information Science, University of Konstanz, Germany. His research interests include machine learning, deep learning, visual quality assessment, and crowd-sourcing.
\end{IEEEbiography}
\vskip 0pt plus -1fil
\begin{IEEEbiography}[{\includegraphics[width=1in,height=1.25in,clip,keepaspectratio]{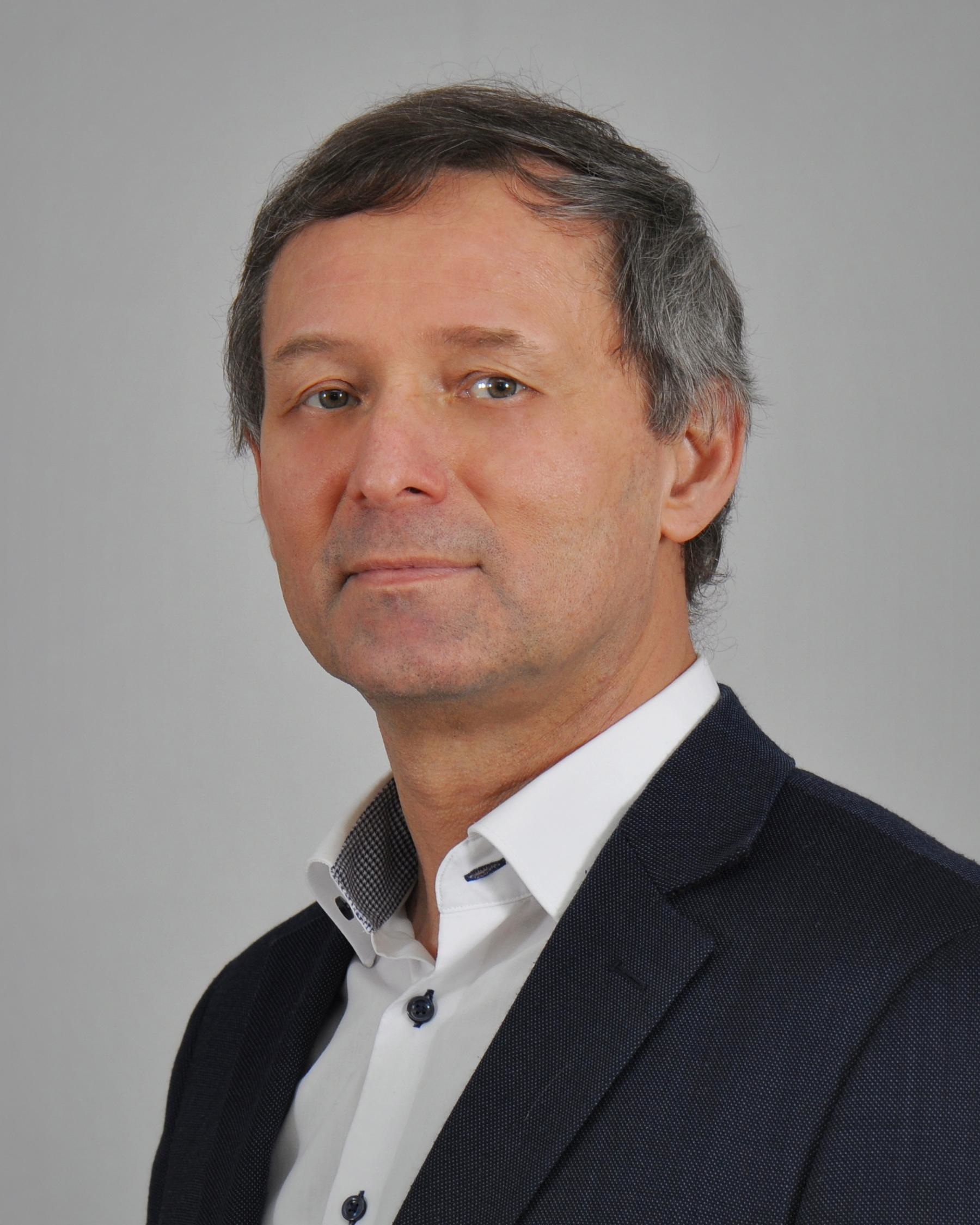}}]{Tamas Sziranyi}received his  Ph.D. and D.Sci. degrees in 1991 and 2001, by the Hungarian Academy of Sciences, Budapest. He was appointed to a Full Professor position in 2001 at Pannon University, Veszprem, Hungary, and, in 2004, at the Peter Pazmany Catholic University, Budapest. Presently he is a full professor at the Budapest University of Technology and Economics. He has been a research scientist at the Institute for Computer Science and Control (SZTAKI) since 1992, where he leads the Machine Perception Research Laboratory since 2006. His research activities include machine perception, pattern recognition, texture and motion segmentation, Markov Random Fields and stochastic optimization, remote sensing, surveillance, intelligent networked sensor systems, graph-based clustering, digital film restoration.  With his research laboratory, he has participated in several prestigious international (ESA, EDA, FP6, FP7, OTKA) projects.

Dr. Sziranyi was the founder and past president (1997 to 2002) of the Hungarian Image Processing and Pattern Recognition Society. He was an Associate Editor of IEEE T. Image Processing (2003-2009),  and he has been an AE of Digital Signal Processing (Elsevier) since 2012 and Remote Sensing (MDPI) from 2019. He was honored by the Master Professor award in 2001, by the Szechenyi professorship and the ProScientia (Pannon University) award in 2011, and by the Officers’Cross by the President of Hungary (2018). He is a Fellow both of the Int. Assoc. Pattern Recognition (IAPR) and the Hungarian Academy of Engineering from 2008. He has more than 300 publications, including 60 in major scientific journals, and several international patents. 
\end{IEEEbiography}
\vskip 0pt plus -1fil
\begin{IEEEbiography}[{\includegraphics[width=1in,height=1.25in,clip,keepaspectratio]{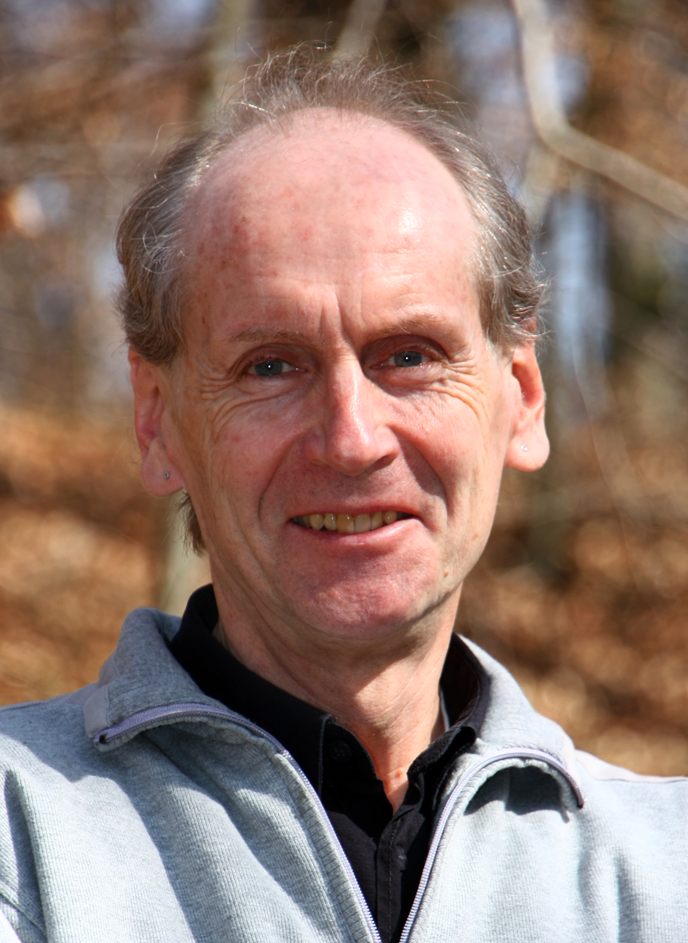}}]{Dietmar Saupe,} since 2002, is a full professor for Computer Science at the University of Konstanz, Germany. He is head of a research group focusing on multimedia signal processing, including applications in sports science. Over the years, his areas of interest include numerical methods, dynamical systems, scientific visualization, computer graphics, image and video compression, medical image processing, computer vision, 3D models, and sports informatics. He has co-authored several award-winning books on fractals and chaos. Currently, his research group is engaged in two projects, one on image and video quality assessment and the other on modeling and optimizing performance in endurance sports. At the University of Konstanz, he has been Chair of the Department of Computer Science and Vice Dean of the Faculty of Natural Sciences for four years. He has been an Editorial Board member of Computer Graphics Forum and the International Journal on Image and Video Processing and was Chief Editor of the Springer Journal Computing (Archives for Scientific Computing).
\end{IEEEbiography}

\newif\ifsupplementary
\end{document}